\documentclass[letterpaper]{article} 
\usepackage{aaai24} 

\usepackage{times}  
\usepackage{helvet}  
\usepackage{courier}  
\usepackage[hyphens]{url}  
\usepackage{graphicx} 
\usepackage{natbib}  
\usepackage{caption} 
\usepackage{algorithm}
\usepackage[noend]{algpseudocode}
\usepackage{newfloat}
\usepackage{listings}
\usepackage{epsfig}
\usepackage{amsmath}
\usepackage{amssymb}
\usepackage{booktabs}
\usepackage{multirow}
\usepackage{amssymb}
\usepackage{pifont}
\usepackage{makecell}
\usepackage{caption}
\usepackage{subcaption}
\usepackage{graphicx}
\usepackage{dsfont}
\usepackage{mathrsfs}
\usepackage{blindtext}
\usepackage[table]{xcolor}
\usepackage{bm}
\usepackage{lipsum}  
\usepackage[english]{babel}
\usepackage[autostyle]{csquotes}
\usepackage[table]{xcolor}
\usepackage[export]{adjustbox}

\definecolor{red}{rgb}{1, 0, 0}

\DeclareCaptionStyle{ruled}{labelfont=normalfont,labelsep=colon,strut=off} 
\floatstyle{ruled}
\newfloat{listing}{tb}{lst}{}
\floatname{listing}{Listing}
\title{FACL-Attack: Frequency-Aware Contrastive Learning\\for Transferable Adversarial Attacks}
\author{
    Hunmin Yang\textsuperscript{\rm 1,\rm 2,}\equalcontrib,
    Jongoh Jeong\textsuperscript{\rm 1,}\equalcontrib,
    Kuk-Jin Yoon\textsuperscript{\rm 1}
}
\affiliations{
    \textsuperscript{\rm 1}Visual Intelligence Lab., KAIST \\
    \textsuperscript{\rm 2}Agency for Defense Development \\

    {\rm \{hmyang, jeong2, kjyoon\}@kaist.ac.kr}
}

\begin{document}
\maketitle

\begin{abstract}
Deep neural networks are known to be vulnerable to security risks due to the inherent transferable nature of adversarial examples.
Despite the success of recent generative model-based attacks demonstrating strong transferability, it still remains a challenge to design an efficient attack strategy in a real-world strict black-box setting, where both the target domain and model architectures are unknown.
In this paper, we seek to explore a feature contrastive approach in the frequency domain to generate adversarial examples that are robust in both cross-domain and cross-model settings.
With that goal in mind, we propose two modules that are only employed during the training phase: a \textbf{F}requency-\textbf{A}ware \textbf{D}omain \textbf{R}andomization (FADR) module to randomize domain-variant low- and high-range frequency components and a \textbf{F}requency-\textbf{A}ugmented \textbf{C}ontrastive \textbf{L}earning (FACL) module to effectively separate domain-invariant mid-frequency features of clean and perturbed image.
We demonstrate strong transferability of our generated adversarial perturbations through extensive cross-domain and cross-model experiments, while keeping the inference time complexity.
\end{abstract}

\section{Introduction}
Deep neural networks have brought forth tremendous improvements in visual recognition tasks.
However, the inherent transferable nature of adversarial examples still exposes the security vulnerability to malicious attackers targeting such susceptible classifiers, causing serious threats and undesirable outcomes in real-world applications.
The majority of current attack methods can be primarily classified into two main categories: iterative or optimization-based approaches, and generative model-based approaches.
Over the past years, iterative attack methods~\cite{goodfellow2014explaining, madry2017towards, croce2020reliable, lorenz2021detecting, dong2018boosting, DI, dr, ssp} have been the standard attack protocol for its simplicity and effectiveness.
However, this iterative approach is frequently constrained by inefficient time complexity and the potential risk of over-fitting to the training data and models.
Moreover, it has shown limited applicability in practical situations due to the low transferability to unknown models and domains.

Regarding the transferability of adversarial attacks, threat model is typically carried out in three different settings (\textit{i.e.}, white-box, black-box, and strict black-box) depending on the prior knowledge of the model architecture and data distributions by the adversary.
In each respective setting, the adversary has either complete knowledge of the target model profile (\textit{i.e.}, architecture and weights) and data distributions reflecting the target domain, query access to the limited black-box only, or no information at all.
In this work, we specifically consider the \textit{strict black-box} case in which the victim attributes are completely \textit{unknown} to the attacker since such a scenario is commonly encountered in practical real-world settings.
We believe that crafting adversarial examples in this strict black box setting has practical values towards stronger transferabilty, as well as safe and reliable deployment of deep learning models.

\begin{figure}
    \centering
    \includegraphics[width=\columnwidth]{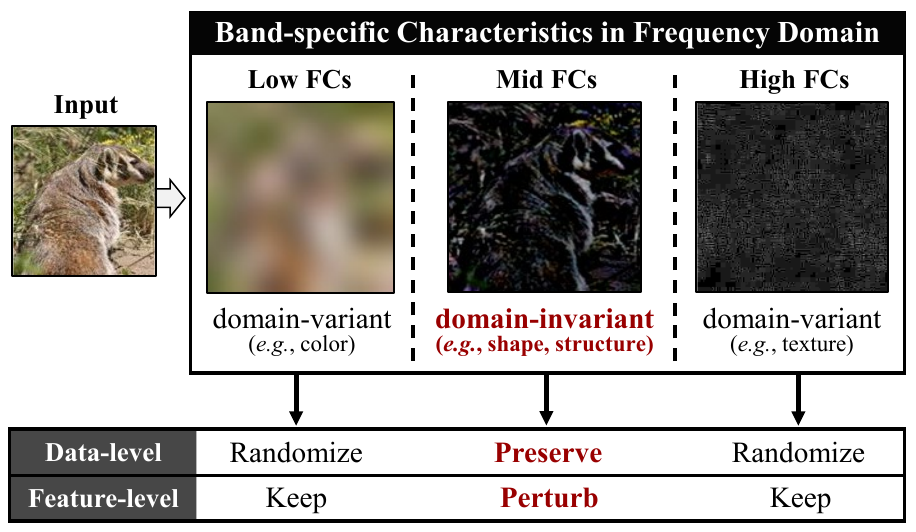}
    \caption{To boost the transferability of adversarial examples, we exploit band-specific characteristics of natural images in the frequency domain. Our method randomizes domain-variant low- and high-band frequency components (FCs) in the data space, and contrasts domain-invariant mid-range clean and perturbed feature pairs in the feature space.}
    \label{fig:motivation}
\end{figure}

In this light, generative attacks~\cite{poursaeed2018generative, naseer2019cross, salzmann2021learning, naseer2021generating, zhang2022beyond} have recently gained attention, demonstrating the high transferability across unknown models and domains.
Moreover, generator-based attacks yield lower time complexity than iterative or optimization-based methods in the inference stage, which is also a crucial part for real-world attacks.
While the current chain of generative attack methods~\cite{poursaeed2018generative, naseer2019cross, naseer2021generating, salzmann2021learning, zhang2022beyond, wu2020boosting} are time-efficient and effective against various black-box settings, we remark that their methods do not actively leverage domain-related characteristics to facilitate more transferable attacks.

In that sense, our work is inspired by frequency domain manipulations~\cite{yin2019fourier, wang2020high, wang2020towards} in domain adaptation (DA)~\cite{yang2020fda} and generalization (DG)~\cite{huang2021fsdr, xu2021fourier}, demonstrating the superior generalization capabilities of the trained model.
As we target transferable attack on unknown target domains and victim models to boost the transferability in a similar setting, we seek to exploit domain-related characteristics from simple yet effective frequency manipulations.

Several recent studies have focused on frequency-based adversarial attacks to manipulate adversarial examples, aimed at deeper understanding of their dataset dependency~\cite{maiya2021frequency}, adversarial robustness~\cite{duan2021advdrop}, and the security vulnerability~\cite{dziugaite2016study}.
With a slightly different motive, SSAH~\cite{luo2022frequency} aims to improve the perceptual quality, whereas \cite{lowfreq} designs low-frequency perturbations to enhance the efficiency of black-box queries.
Although low-frequency perturbations are efficient, they are known to provide less effective transfer between models~\cite{lowfreqeffective}.
As such, we delve deeper into frequency-driven approaches that effectively enhance the transferability of adversarial examples, especially crafted in a generative framework.

To this end, we propose a novel generative attack method, \textbf{FACL-Attack}, to facilitate transferable attacks across various domains and models from the frequency domain perspective.
In our training, we introduce frequency-aware domain randomization and feature contrastive learning, explicitly leveraging band-specific characteristics of image attributes such as color, shape, and texture, as illustrated in Figure~\ref{fig:motivation}.
We highlight our contributions as follows:
\begin{itemize}
    \item We propose two modules to boost the adversarial transferability, \textit{FADR} and \textit{FACL}, in which FADR randomizes \textit{domain-variant} data components while FACL contrasts \textit{domain-invariant} feature pairs in the frequency domain.
    \item We achieve the state-of-the-art attack transferability across various domains and model architectures, demonstrating the effectiveness of our method.
    \item Our plug-and-play modules can be easily integrated into existing generative attack frameworks, further boosting the transferability while keeping the time complexity.
\end{itemize}

\begin{figure*}[!t]
    \centering
    \includegraphics[width=\textwidth]{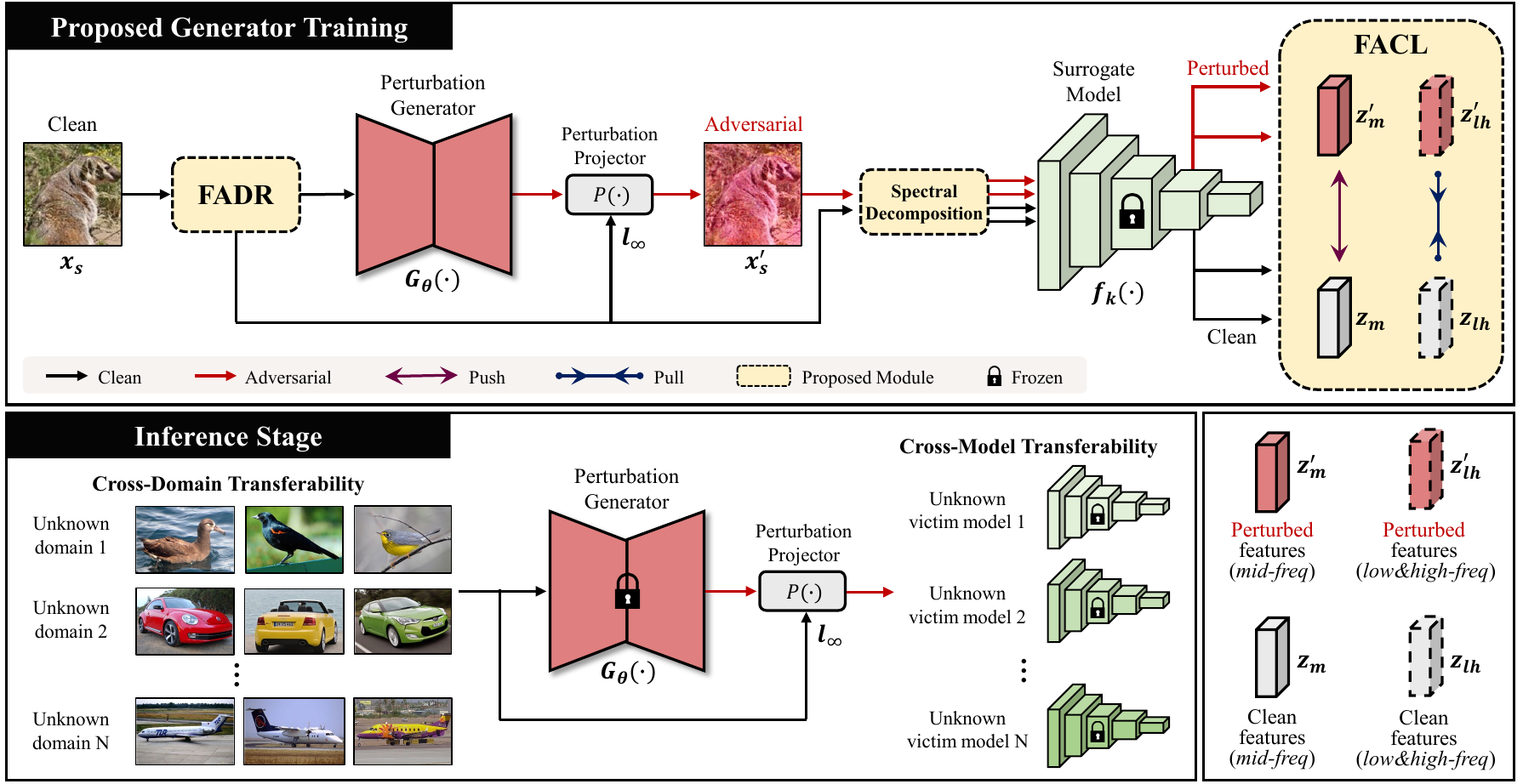}
    \caption{Overview of FACL-Attack.
    From the clean input image, our FADR module outputs the augmented image after spectral transformation, which is targeted to randomize only the domain-variant low/high FCs.
    The perturbation generator $G_{\theta}(\cdot)$ then produces the $l_{\infty}$-budget bounded adversarial image $\boldsymbol{\mathit{x}}'_{s}$ with perturbation projector $P(\cdot)$ from the randomized image.
    The resulting clean and adversarial image pairs are decomposed into mid-band (domain-agnostic) and low/high-band (domain-specific) FCs, whose features $f_{k}(\cdot)$ extracted from the $k$-th layer of the surrogate model are contrasted in our FACL module to boost the adversarial transferability.
    The adversarial image $\boldsymbol{\mathit{x}}'_{s}$ is colorized only for visualization.}
    \label{fig:overview}
\end{figure*}

\section{Related Work}
\subsection{Generator-based Adversarial Attack}
Generative attack~\cite{poursaeed2018generative} employs the concept of adversarial training~\cite{goodfellow2020generative} to create perturbations across entire data distributions.
This is achieved by regarding a pre-trained surrogate model as a discriminator, and it is advantageous due to the ability of generating diverse forms of perturbations across multiple images simultaneously.
Existing methods aim to enhance the generator training by leveraging both the cross-entropy (CE) loss~\cite{poursaeed2018generative} and the relativistic CE loss~\cite{naseer2019cross}, improving the transferability across domains and models.
Recent studies~\cite{salzmann2021learning, zhang2022beyond} utilize features extracted from the mid-level layers of the surrogate model, which encompass a higher degree of shared information among different model architectures.
We follow the traces of the recent works and explore a method to further enhance the transferability by introducing a novel perspective from the frequency domain.

\subsection{Frequency-based Approach for Generalization}
Convolutional neural networks are known to exhibit intriguing attributes within the frequency domain~\cite{yin2019fourier, tsuzuku2019structural, yin2019fourier, wang2020high, wang2020towards}, demonstrating proficient generalization capability by effectively harnessing the band-specific information derived from Fourier filtering~\cite{dziugaite2016study, guo2017countering, long2022frequency}.
Spectral manipulations for enhancing the generalization capability can be achieved through simple yet powerful transformations like the Fast Fourier Transform (FFT), which dissects an image into amplitude components that vary across domains and phase components that remain consistent across different domains~\cite{xu2021fourier}.
The Discrete Cosine Transform (DCT) also serves as an efficient technique to decompose spectral elements into domain-agnostic mid-frequency components (mid-FCs) and domain-specific low- and high-FCs, which contributed to the effective spectral domain randomization in FSDR~\cite{huang2021fsdr}.
In our work, we also employ the DCT to decompose images into domain-agnostic and domain-specific frequency components, facilitating the effective domain randomization and feature-level contrastive learning for transferable attacks.

\subsection{Feature Constrastive Learning}
Manipulating image representations in the feature space has demonstrated significant performance improvement in real-world scenarios characterized by domain shifts.
In the field of DA and DG, common approaches such as feature alignment~\cite{yang2022contrastive} and intra-class feature distance minimization with inter-class maximization~\cite{kang2019contrastive, luo2022frequency, doublecontrast2022} are successful in mitigating the domain discrepancies.
Specifically, several studies~\cite{wang2022cross, kim2021selfreg} have directly addressed the domain gap issue by manipulating pairs of domain-invariant representations from various domains that correspond to samples of the same class.
Continuing in the realm of generative attacks, recent studies have employed CLIP-based~\cite{aich2022gama} and object-centric~\cite{aich2023leveraging} features for effective training of the perturbation generator.
In our work, we leverage frequency-augmented feature contrastive learning on domain-agnostic mid-band feature pairs.
Simultaneously, we reduce the significance of domain-specific features in the low- and high-bands to improve the adversarial transferability.

\section{Proposed Attack Method: FACL-Attack}
\label{sec:our_method}
\DeclareRobustCommand{\rchi}{{\mathpalette\irchi\relax}}
\newcommand{\irchi}[2]{\raisebox{\depth}{$#1\chi$}} 

\subsubsection{Problem definition.}
Generating adversarial examples revolves around solving an optimization problem, whereas generating transferable adversarial examples addresses the challenge of generalization.
Our goal is to train a generative model $G_{\theta}(\cdot)$ to craft adversarial perturbations $\delta$ that are well transferable to arbitrary domains and victim models aimed to trigger mispredictions on the image classifier $f(\cdot)$.
Specifically, the generator maps the clean image $\boldsymbol{\mathit{x}}$ to its corresponding adversarial example $\boldsymbol{\mathit{x}}'=G_{\theta}(\boldsymbol{\mathit{x}})$ containing perturbations constrained by $\|\delta\|_{\infty}\leq\epsilon$.

\subsubsection{Overview of FACL-Attack.}
Our method aims to train a robust perturbation generator that yields effective adversarial examples given arbitrary images from black-box domains to induce the unknown victim model to output misclassification.
It consists of two key modular operations in the frequency domain, each applied to the input image data and features extracted from the surrogate model only during the training stage, as illustrated in Figure~\ref{fig:overview}.

As inspired by the power of frequency domain augmentation in domain generalization~\cite{huang2021fsdr, xu2021fourier}, our first module, \textit{\textbf{F}requency-\textbf{A}ware \textbf{D}omain \textbf{R}andomization (FADR)}, transforms a pixel-domain image to the frequency-domain components using DCT.
It randomizes \textit{domain-variant} low- and high-frequency band components and preserves \textit{domain-invariant} mid-frequency components in the input image.
Then a perturbation generator is trained to craft bounded adversarial images $\boldsymbol{\mathit{x}}'_{s}$, \textit{i.e.}, perturbation $\delta$ added to the clean image $\boldsymbol{\mathit{x}}_{s}$ and constrained by perturbation projector $P(\cdot)$.
We then spectrally decompose the randomized $\boldsymbol{\mathit{x}}_{s}$ and $\boldsymbol{\mathit{x}}'_{s}$ into each low- and high-band, and mid-band frequency component, which are inversely transformed to the image domain by IDCT and passed through the pre-defined surrogate model for feature extraction.
Following the recent line of works~\cite{salzmann2021learning, zhang2022beyond} on transferable generative attacks, we leverage the mid-layer features $f_{k}(\cdot)$ for feature contrastive learning.
Each band-specific clean and perturbed feature pair is contrasted in our \textit{\textbf{F}requency-\textbf{A}ugmented \textbf{C}ontrastive \textbf{L}earning (FACL)} module, whereby \textit{domain-agnostic} mid-band FC pair is to repel and \textit{domain-specific} low- and high-band FC pair is to attract each other. This straightforward but effective data- and feature-level guidance in the frequency domain significantly contributes to boost the adversarial transferability as demonstrated in the following sections.

\begin{figure}[!t]
\centering
\includegraphics[width=\linewidth]{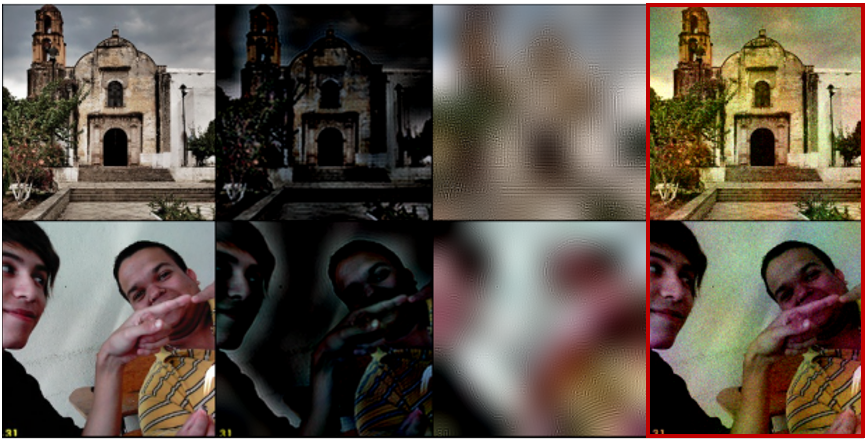}
\caption{Visualization of spectral transformation in FADR.
From the clean input image (column 1), our FADR decomposes the image into mid-band (column 2) and low/high-band (column 3) FCs.
The FADR only randomizes the low/high-band FCs, yielding the augmented output in column 4.
Here we demonstrate transformations with large hyper-parameters of $\rho=0.5$ and $\sigma=8$ for visualization.
}
\label{fig:FADR}
\end{figure}

\subsection{Frequency-Aware Domain Randomization}
\label{sec:fadr}
This subsection describes our FADR module designed to boost the robustness of perturbation generator $G_{\theta}(\cdot)$ against arbitrary domain shifts in practical real-world scenarios.
Inspired by recent works that convert the training image from pixel space into frequency space for boosting the domain generalization capabilities~\cite{huang2021fsdr, xu2021fourier}, we decompose the input training images into multiple-range FCs by leveraging DCT, and apply random masked filtering operation on \textit{domain-specific} image attributes that lie in the low- and high-frequency bands.
While FSDR~\cite{huang2021fsdr} and FACT~\cite{xu2021fourier} each employs histogram matching and Fourier-based amplitude mix-up, our proposed FADR module explicitly manipulates the DCT coefficients to diversify input images, aligning with a recent work~\cite{long2022frequency} that narrows the gap between the surrogate model and possible victim models via spectrum transformation.
We remark that our approach applies domain randomization exclusively to \textit{domain-specific} FCs that are subject to change from various domains, whereas the existing work~\cite{long2022frequency} applies spectral transformation over the whole frequency bands containing not only domain-specific information, but also domain-agnostic semantic details.

In converting the input images into the frequency domain, we apply DCT to each channel separately.
We then apply random masked filtering to diversify the input images for boosting the cross-domain transferability.
Our spectral transformation operation $\mathcal{T}_{\mathrm{FADR}}(\cdot)$ for source images $\boldsymbol{\mathit{x}}_{s}$ can be mathematically expressed as follows:
\begin{equation}
    \begin{aligned}
        \mathcal{T}_{\mathrm{FADR}}(\boldsymbol{x}_{s}) & = \phi^{-1}\Big((\phi(\boldsymbol{x}_{s}+\boldsymbol{\xi})\odot \boldsymbol{\mathit{M}}\Big),
    \end{aligned}
\label{eq:T_FADR}
\end{equation}

\noindent with the mask $\boldsymbol{\mathit{M}}$ defined as follows:
\begin{equation}
    \boldsymbol{\mathit{M}} =
    \begin{cases} 
      \mathcal{U}(1-\rho,1+\rho), & \mathrm{if}~f<f_l, \\
      1, & \mathrm{if}~f_l \leq f < f_h, \\
      \mathcal{U}(1-\rho,1+\rho), & \mathrm{if}~f \geq f_h,
    \end{cases}
    \label{eq:mask}
\end{equation}
\noindent where $\odot$, $\phi$, $\phi^{-1}$ denote Hadamard product, DCT, and inverse DCT (IDCT) operation, respectively.
The random noise $\boldsymbol{\xi}\sim\mathcal{N}(0,\sigma^{2}\mathbf{I})$ is sampled from a Gaussian distribution, and the mask values are randomly sampled from Uniform distribution, denoted as $\mathcal{U}$.
For the random mask matrix $\boldsymbol{\mathit{M}}$ which has same dimension with the DCT output, we assign its matrix component values as defined in Equation~\ref{eq:mask}, where we set the low and high thresholds as $f_l$, and $f_h$, respectively, to distinguish low-, mid-, and high-frequency bands.
Note that we can parameterize our FADR module with hyper-parameters $\rho$ and $\sigma$.
The spectral transformation in our FADR module is conceptually illustrated in Figure~\ref{fig:FADR}.

The augmented image output from FADR is then fed as input to the generator $G_{\theta}(\cdot)$ to yield the adversarial image $\boldsymbol{\mathit{x}}'_{s} = P(G_{\theta}(\mathcal{T}_{\mathrm{FADR}}(\boldsymbol{\mathit{x}_{s}})))$, after the perturbation projection within the pre-defined budget of $\|\delta\|_{\infty}\leq\epsilon$. 

\subsection{Frequency-Augmented Contrastive Learning}
\label{sec:facl}
Recent works on multi-object scene attacks have highlighted the importance of feature-level contrast for transferable generative attacks.
In a similar approach to their ideas of exploiting local patch differences~\cite{aich2023leveraging} or CLIP features~\cite{aich2022gama}, our FACL module seeks to apply feature contrast specifically in the \textit{domain-agnostic} mid-frequency range for improving the generalization capability of the trained perturbation generator $G_{\theta}(\cdot)$. 

\subsubsection{Spectral decomposition.} According to the training pipeline of our FACL-Attack in Figure~\ref{fig:overview}, the generated adversarial image $\boldsymbol{\mathit{x}}'_{s}$ undergoes spectral decomposition before feature extraction from the surrogate model.
This process is carried out by using a band-pass filter $\boldsymbol{\mathit{M}}_{\mathrm{bp}}$ and a band-reject filter $\boldsymbol{\mathit{M}}_{\mathrm{br}}$, to decompose the surrogate model inputs into mid- and low/high-band FCs, respectively.
The spectral decomposition operator is defined as follows:
\begin{equation}
    \boldsymbol{\mathit{M}}_{\mathrm{bp}}=
    \begin{cases} 
      1, & \mathrm{if}~f_l \leq f < f_h, \\
      0, & \mathrm{otherwise}, \\
    \end{cases}  
    \label{eq:M_bp}
\end{equation}
\noindent where $\boldsymbol{\mathit{M}}_{\mathrm{br}}$ is the opposite of $\boldsymbol{\mathit{M}}_{\mathrm{bp}}$, holding its values in reverse.
Then the spectrally decomposed features from the surrogate model $\boldsymbol{\mathit{f}}$ are defined as:
\begin{equation}
    \mathbf{z}_{\mathrm{band}} = \mathit{f}_{k}\Big(\phi^{-1}\Big(\phi(\boldsymbol{\mathit{x}}_{\mathrm{input}})\odot \boldsymbol{\mathit{M}}_{\mathrm{band}}\Big)\Big),
    \label{eq:spectrally_decomposed_features}
\end{equation}
\noindent where $\boldsymbol{\mathit{M}}_{\mathrm{band}}$ is set to either $\boldsymbol{\mathit{M}}_{\mathrm{bp}}$ or $\boldsymbol{\mathit{M}}_{\mathrm{br}}$, and $\mathit{f}_{k}(\cdot)$ denotes the $k$-th layer of $\mathit{f}$.
Given $\boldsymbol{\mathit{x}}_{s}$ and $\boldsymbol{\mathit{x}}'_{s}$ as input, we finally obtain two pairs of band-specific frequency-augmented features to contrast, \textit{i.e.}, $(\mathbf{z}_{m}, \mathbf{z}'_{m})$ for repelling, and $(\mathbf{z}_{lh}, \mathbf{z}'_{lh})$ for attracting each other.

\subsubsection{Loss function.} 
\begin{figure}[!t]
\centering
\includegraphics[width=\linewidth]{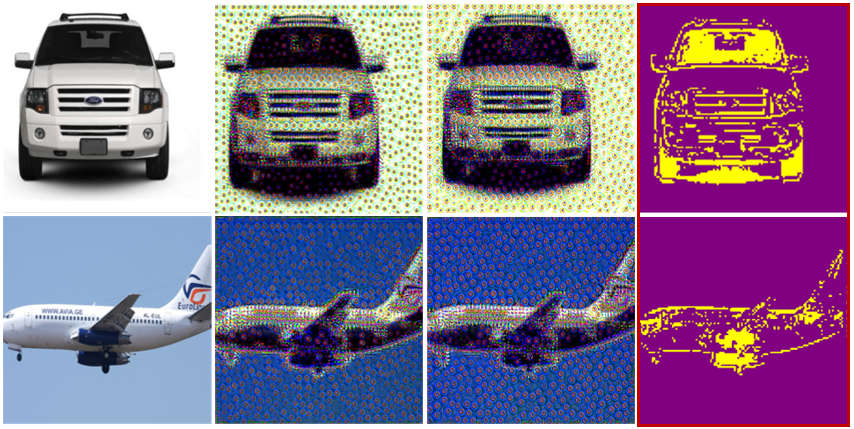}
\caption{Clean image, unbounded adversarial images from baseline and FACL, and the final difference map (Diff(baseline, baseline+FACL)), from left to right.
Our generated adversarial perturbations are more focused on domain-agnostic semantic region such as shape, facilitating more transferable attack.}
\label{fig:diff_map}
\end{figure}

The baseline loss $\mathcal{L}_{\mathrm{orig}}$ for attacking the surrogate model via contrasting clean and adversarial feature pairs is defined as follows:
\begin{equation}
    \mathcal{L}_{\mathrm{orig}}=\mathrm{sim}(\mathit{f}_{k}(\boldsymbol{\mathit{x}}_{s}), \mathit{f}_{k}(\boldsymbol{\mathit{x}}'_{s})),
\label{eq:loss_original}
\end{equation}
\noindent where $\mathrm{sim}$ refers to the standard cosine similarity metric.
To boost the attack transferability, our FACL module effectively exploits the spectrally decomposed feature pairs in our proposed FACL loss function defined as follows:
\begin{equation}
        \mathcal{L}_{\mathrm{FACL}} =  \mathrm{sim}(\mathbf{z}_{m}, \mathbf{z}'_{m}) - \mathrm{sim}(\mathbf{z}_{lh}, \mathbf{z}'_{lh}),
\label{eq:loss_FACL}
\end{equation}
\noindent where the goal of $\mathcal{L}_{\mathrm{FACL}}$ is to reinforce the effectiveness of \textit{domain-agnostic} mid-band feature contrast ($\mathbf{z}_{m}, \mathbf{z}'_{m}$), while minimizing the importance of \textit{domain-specific} low- and high-band feature difference ($\mathbf{z}_{lh}, \mathbf{z}'_{lh}$).
In this approach, our $\mathcal{L}_{\mathrm{FACL}}$ facilitates the push-pull action among the band-specific feature pairs, further guiding the perturbation generation towards more robust regime, as shown in Figure~\ref{fig:diff_map}.

\subsubsection{Final learning objective.} 
We train our perturbation generator by minimizing the total loss function as follows:
\begin{equation}
    \min_{\theta}\; (\lambda_{\mathrm{orig}}\cdot\mathcal{L}_{\mathrm{orig}} + \lambda_{\mathrm{FACL}}\cdot\mathcal{L}_{\mathrm{FACL}}),
\label{eq:final_objective}
\end{equation}
\noindent
where $\lambda_{\mathrm{orig}}$ and $\lambda_{\mathrm{FACL}}$ are loss coefficients.
The objective guides our generator $G_{\theta}(\cdot)$ to generate more robust perturbations to domain shifts as well as model variances.

\section{Experiments}
\subsection{Experimental Setup}
\subsubsection{Datasets and attack settings.}
We evaluate our method over challenging strict black-box settings (\textit{i.e.}, \textit{cross-domain} and \textit{cross-model}) in the image classification task.
We set the target domain and victim model to be different from the source domain and surrogate model.
The perturbation generator is trained on ImageNet-1K~\cite{imagenet} and evaluated on CUB-201-2011~\cite{cub}, Stanford Cars~\cite{car}, and FGVC Aircraft~\cite{air}.
As BIA~\cite{zhang2022beyond} highlights the importance of using a large-scale dataset for training, we train on ImageNet-1K accordingly.
For the cross-model setting, we evaluate our method over black-box models but white-box domain (\textit{i.e.}, ImageNet-1K) setting.
The details for the datasets are described in Table~\ref{tab:datasets}.

\subsubsection{Surrogate and victim models.}
Our perturbation generator is trained against ImageNet-1K pre-trained surrogate models (\textit{e.g.}, VGG-16~\cite{vgg}).
For the cross-model evaluation, we investigate other architectures including VGG-19~\cite{vgg}, ResNet50 (Res-50), ResNet152 (Res-152)~\cite{res152}, DenseNet121 (Dense-121), DenseNet169 (Dense-169)~\cite{densenet} and Inception-v3 (Inc-v3)~\cite{inc-v3}.
For the cross-domain setting (\textit{i.e.}, CUB-201-2011, Stanford Cars, and FGVC Aircraft), we use fine-grained classification models trained under DCL framework~\cite{dcl} with three different backbones, which include Res-50, SENet154 and SE-ResNet101 (SE-Res101)~\cite{senet}.

\begin{table}[!t]
\fontsize{10pt}{10pt}\selectfont
\centering
\resizebox{\linewidth}{!}{
    \begin{tabular}{lccc}
        \toprule
        
        \textbf{Dataset} & \textbf{$\#$ Class} & \textbf{$\#$ Train / Val.} & \textbf{Resolution} \\

        \midrule
        \midrule
         
        ImageNet-1K & 1,000 & 1.28~M / 50,000 & 224$\times$224 \\
        CUB-200-2011 & 200 & 5,994 / 5,794 & 448$\times$448 \\
        Stanford Cars & 196 & 8,144 / 8,041 & 448$\times$448 \\
        FGVC Aircraft &  100 & 6,667 / 3,333 & 448$\times$448 \\
        
        \bottomrule                     
    \end{tabular}
    }
\caption{Description of datasets.}
\label{tab:datasets}
\end{table}

\subsubsection{Implementation details.}
We closely follow the implementation of recent works on generative attacks~\cite{naseer2019cross, salzmann2021learning, zhang2022beyond, aich2022gama} for fair comparison.
Our perturbation generator consists of down-sampling, residual, and up-sampling blocks that translate clean images into adversarial examples.
The surrogate model layer from which we extract features is \textit{Maxpool}.3 for VGG-16.
We train with an Adam optimizer ($\beta_{1}=0.5$, $\beta_{2}=0.999$)~\cite{adam} with the learning rate  of $2\times10^{-4}$, and the batch size of $16$ for $1$ epoch.
The perturbation budget for crafting the adversarial image is $l_\infty \leq 10$.
For the FADR hyper-parameters, we follow a prior work~\cite{huang2021fsdr} to set the low and high frequency threshold to $f_l=7$ and $f_h=112$, respectively.
We use $\rho=0.01$ and $\sigma=8$ for spectral transformation and describe more details in \textit{Supplementary}.

\begin{table*}[!t]
\fontsize{10pt}{10pt}\selectfont
\setlength{\tabcolsep}{1.5pt}
\centering
\resizebox{\linewidth}{!}{
    \begin{tabular}{ccccccccccccc}
        \toprule
        \multirow{2}{*}{\textbf{Method}} & \multicolumn{3}{c}{\textbf{CUB-200-2011}} && \multicolumn{3}{c}{\textbf{Stanford Cars}} && \multicolumn{3}{c}{\textbf{FGVC Aircraft}} & \multirow{2}{*}{\textbf{AVG.}} \\
        \cmidrule{2-12}
         
        & \textbf{Res-50} & \textbf{SENet154} & \textbf{SE-Res101} && \textbf{Res-50} & \textbf{SENet154} & \textbf{SE-Res101} &&  \textbf{Res-50} & \textbf{SENet154} & \textbf{SE-Res101} \\
        \midrule
        \midrule
        
        Clean & 87.35 & 86.81 & 86.56 && 94.35 & 93.36 & 92.97 && 92.23 & 92.08 & 91.90 & 90.85 \\
        \midrule

        GAP~\cite{poursaeed2018generative} & 68.85 & 74.11 & 72.73 && 85.64 & 84.34 & 87.84 && 81.40 & 81.88 & 76.90 & 79.30 \\
        
        CDA~\cite{naseer2019cross} & 69.69 & 62.51 & 71.00 && 75.94 & 72.45 & 84.64 && 71.53 & \underline{58.33} & 63.39 &69.94 \\
        
        LTP~\cite{salzmann2021learning} & \underline{30.86} & \underline{52.50} & 62.86 && \underline{34.54} & \textbf{65.53} & 73.88 && \textbf{15.90} & 60.37 & 52.75 & \underline{49.91} \\
        
        BIA~\cite{zhang2022beyond} & 32.74 & 52.99 & \underline{58.04} && 39.61 & 69.90 & \underline{70.17} && 28.92 & 60.31 & \textbf{46.92} & 51.07 \\
        
        \midrule    
        
        \textbf{FACL-Attack~(Ours)} & \textbf{24.74} & \textbf{44.06} & \textbf{53.75} && \textbf{26.58} & \underline{65.71} & \textbf{61.40} && \underline{19.72} & \textbf{52.01} & \underline{48.51} & \textbf{44.05} \\                                     
        \bottomrule
    \end{tabular}
}
\caption{Cross-domain evaluation results. The perturbation generator is trained on ImageNet-1K with VGG-16 as the surrogate model and evaluated on black-box domains with black-box models.
We compare the top-1 classification accuracy after attacks with the perturbation budget of $l_\infty \leq 10$ (the lower, the better).
}
\label{tab:cross_domain}
\end{table*}

\begin{table*}[!t]
\fontsize{10pt}{10pt}\selectfont
\setlength{\tabcolsep}{4.5pt}
\centering
\resizebox{\linewidth}{!}{
    \begin{tabular}{ccccccccccccccccccc}
        \toprule 
        \textbf{Method} &
        \textbf{Venue} &
        \multicolumn{1}{c}{\textbf{VGG-16}} &
        \multicolumn{1}{c}{\textbf{VGG-19}} &
        \multicolumn{1}{c}{\textbf{Res-50}} &
        \multicolumn{1}{c}{\textbf{Res-152}} &
        \multicolumn{1}{c}{\textbf{Dense-121}} &
        \multicolumn{1}{c}{\textbf{Dense-169}} &
        \multicolumn{1}{c}{\textbf{Inc-v3}} &                
        \multirow{1}{*}{\textbf{AVG.}} \\
        \midrule
        \midrule
        
        Clean & - & 70.14 & 70.95 & 74.61 & 77.34 & 74.22 & 75.75 & 76.19 & 74.17 \\         
        \midrule
         
        GAP~\cite{poursaeed2018generative} & CVPR'18 & 23.63 & 28.56 & 57.87 & 65.50 & 57.94 & 61.37 & 63.30 & 55.76 \\
          
        CDA~\cite{naseer2019cross} & NeurIPS'19 & \textbf{0.40} & \textbf{0.77} & 36.27 & 51.05 & 38.89 & 42.67 &  54.02 & 32.01 \\
        
        LTP~\cite{salzmann2021learning} & NeurIPS'21 & 1.61  & \underline{2.74}  & \underline{21.70} & \underline{39.88} & \underline{23.42} & \textbf{25.46} & 41.27 & \underline{22.30} \\
        
        BIA~\cite{zhang2022beyond} & ICLR'22 & 1.55 &  3.61 & 25.36 & 42.98 & 26.97 & 32.35 & \underline{41.20} & 24.86 \\
        
        \midrule
        
        \textbf{FACL-Attack~(Ours)} & - & \underline{1.45} &  2.92 & \textbf{19.72} & \textbf{36.61} & \textbf{21.34} & \underline{25.61} & \textbf{29.97} & \textbf{19.66} \\

        \bottomrule
         
    \end{tabular}
}
\caption{Cross-model evaluation results. The perturbation generator is trained on ImageNet-1K with VGG-16 as the surrogate model and evaluated on black-box models including white-box model (\textit{i.e.}, VGG-16).
We compare the top-1 classification accuracy after attacks with the perturbation budget of $l_\infty \leq 10$ (the lower, the better).
}
\label{tab:cross_model}
\end{table*}

\subsubsection{Evaluation metric and competitors.}
We choose the top-1 classification accuracy after attacks as our main evaluation metric, unless otherwise stated.
The reported results are the average values obtained from three random seed runs.
The competitors include the state-of-the-art generative attacks such as  GAP~\cite{poursaeed2018generative}, CDA~\cite{naseer2019cross}, LTP~\cite{salzmann2021learning}, and BIA~\cite{zhang2022beyond}.
We set BIA as our baseline.

\subsection{Main Results}
\subsubsection{Cross-domain evaluation results.}
We compare our FACL-Attack with the state-of-the-art generative-model based attacks on various black-box domains with black-box models.
During the training stage, we leverage the ImageNet-1K as the source domain to train a perturbation generator against a pre-trained surrogate model.
In the inference stage, the trained perturbation generator is evaluated on various black-box domains (\textit{i.e.}, CUB-200-2011, Stanford Cars, and FGVC Aircraft) with black-box victim models.
The victim models include pre-trained models which were trained via DCL framework with three different backbones (\textit{i.e.}, Res-50, SENet154, and SE-Res101).

As shown in Table~\ref{tab:cross_domain}, our FACL-Attack outperforms on most \textit{cross-domain} benchmarks, among which are also \textit{cross-model}, by significant margins.
This demonstrates the strong and robust transferable capability of the generator trained by our novel approach with data- and feature-level guidance in the frequency domain.
We posit that the remarkable generalization ability of FACL-Attack owes to the synergy between our two proposed modules that effectively guide feature-level separation in the \textit{domain-agnostic} mid-frequency band (\textit{i.e.}, FACL), complemented by data-level randomization only applied to the \textit{domain-specific} frequency components (\textit{i.e.}, FADR).
In other words, our spectral approach does help improve the generalization capability of the perturbation generator to other black-box domains as well as unknown network architectures.
Moreover, our proposed training modules are complementary with existing generative attack frameworks and can further improve the attack transferability, as shown in \textit{Supplementary}.

\subsubsection{Cross-model evaluation results.}
Although we demonstrated the effectiveness of FACL-attack on boosting the transferability in strict black-box settings (\textit{i.e.}, \textit{cross-domain} as well as \textit{cross-model}) as shown in Table~\ref{tab:cross_domain}, we further investigated on the black-box model scenario in a controlled white-box domain (\textit{i.e.}, ImageNet-1K).
In other words, the generator is trained against a surrogate model (\textit{i.e.}, VGG-16) and evaluated on various victim models which include VGG-16 (white-box), VGG-19, Res-50, Res-152, Dense-121, Dense-169, and Inc-v3.

As shown in Table~\ref{tab:cross_model}, ours also outperforms on most generative attacks where they seem to partially overfit to the white-box model (\textit{i.e.}, VGG-16).
Our outperforming results validate the strong transferability in \textit{cross-model} attacks, in addition to \textit{cross-domain}. 
We posit that the frequency-augmented feature learning could help the perturbation generator craft more robust perturbations, which exhibit better generalization capability to unknown feature space.
This aligns with a recent finding~\cite{long2022frequency} that spectral data randomization contributes to enhance the transferability via simulating diverse victim models.

\begin{table}[!t]
\fontsize{10pt}{10pt}\selectfont
\centering
\setlength{\tabcolsep}{2pt} 
\resizebox{\columnwidth}{!}{
\begin{tabular}{cccccc}
\toprule

\multicolumn{1}{c|}{\multirow{1}{*}{Method}} &
\multirow{1}{*}{$\mathcal{L}_{\mathrm{orig}}$} &
\multicolumn{1}{c}{\multirow{1}{*}{$\mathcal{T}_{\mathrm{FADR}}$}} &
\multicolumn{1}{c|}{\multirow{1}{*}{$\mathcal{L}_{\mathrm{FACL}}$}} &

\multicolumn{1}{c}{\small{Cross-Domain}} & \multicolumn{1}{c}{\small{Cross-Model}} \\

\midrule
\midrule

\multicolumn{1}{c|}{Clean} &  &  & \multicolumn{1}{c|}{} & 90.85 & 74.17 \\
\midrule

\multicolumn{1}{c|}{Baseline} & \multicolumn{1}{c}{\checkmark} & & \multicolumn{1}{c|}{} & 51.07 & 24.86 \\

\multicolumn{1}{c|}{FADR} & \multicolumn{1}{c}{\checkmark} & \multicolumn{1}{c}{\checkmark} & \multicolumn{1}{c|}{} & 46.24 & 20.28 \\

\multicolumn{1}{c|}{FACL} & \multicolumn{1}{c}{\checkmark} & & \multicolumn{1}{c|}{\checkmark} & 45.36 & 20.70 \\

\multicolumn{1}{c|}{\textbf{Ours}} & \multicolumn{1}{c}{\checkmark} & \multicolumn{1}{c}{\checkmark} & \multicolumn{1}{c|}{\checkmark} & \textbf{44.05} & \textbf{19.66} \\
\bottomrule
\end{tabular}
}
\caption{Ablation study on our proposed modules. $\mathcal{T}_{\mathrm{FADR}}$ and $\mathcal{L}_{\mathrm{FACL}}$ are defined in Eq.~\ref{eq:T_FADR} and \ref{eq:loss_FACL}, respectively.}
\label{tab:ablation_modules}
\end{table}

\subsection{More Analyses}
\subsubsection{Ablation study on our proposed modules.}
\label{sec:ablation}
We examined different attack designs to find out how our proposed modules contribute to the attack transferability.
As shown in Table~\ref{tab:ablation_modules}, we trained the perturbation generator by employing each method and evaluated under realistic black-box settings.
\textit{Cross-Domain} is defined as ImageNet-1K $\rightarrow$ \{CUB-200-2011, Stanford Cars, FGVC Aircraft\} and \textit{Cross-Model} indicates VGG-16 $\rightarrow$ \{VGG-16, VGG-19, Res-50, Res-152, Dense-121, Dense-169, Inc-v3\}.
\textit{Baseline} is trained with $\mathcal{L}_\mathrm{orig}$ without any data randomization or band-specific feature contrast.
\textit{FADR} is trained with $\mathcal{L}_\mathrm{orig}$ and frequency-aware domain randomization using $\mathcal{T}_\mathrm{FADR}$. 
\textit{FACL} is trained with $\mathcal{L}_\mathrm{orig}$ and band-specific feature contrast using $\mathcal{L}_\mathrm{FACL}$.

As shown in Table~\ref{tab:ablation_modules}, \textit{Baseline} trained with naive mid-layer feature contrast (\textit{i.e.}, $\mathcal{L}_\mathrm{orig}$) does not perform well due to the domain bias and model over-fitting. \textit{FADR} and \textit{FACL} each outperforms \textit{Baseline} by a large margin, demonstrating the importance of selectively randomizing the \textit{domain-variant} data components and contrasting \textit{domain-invariant} feature pairs for boosting the black-box transferability, respectively.
Furthermore, \textit{Ours} performs the best consistently.
We speculate that \textit{FADR} and \textit{FACL} are complementary since data augmentation through our FADR facilitates the stable feature contrastive learning.

\begin{table}[!t]
\fontsize{10pt}{10pt}\selectfont
\centering
\setlength{\tabcolsep}{7pt} 
\resizebox{\columnwidth}{!}{
\begin{tabular}{c|cccc}
\toprule
Method & Clean & Baseline & All-Rand & \textbf{Ours} \\
\midrule
\midrule
Cross-Domain & 90.85 & 51.07 & 47.24 & \textbf{44.05} \\
Cross-Model & 74.17 & 24.86 & 21.68 & \textbf{19.66} \\
\bottomrule
\end{tabular}
}
\caption{Comparison with domain randomization on the entire frequency band.}
\label{tab:freq_range_comparisons}
\end{table}

\begin{table}[!t]
\fontsize{10pt}{10pt}\selectfont
\centering
\setlength{\tabcolsep}{4pt} 
\resizebox{\columnwidth}{!}{
\begin{tabular}{l|c|ccc}
\toprule
\multicolumn{1}{c|}{Method} & Accuracy~$\downarrow$ & SSIM$\uparrow$ & PSNR$\uparrow$ & LPIPS$\downarrow$ \\
\midrule 
\midrule
BIA ($l_{\infty}\le10$) & 24.86 & 0.73 & 28.71 & 0.49 \\ 
\textbf{Ours} ($l_{\infty}\le10$) & \textbf{19.66} & 0.72 & 28.61 & 0.49 \\
\midrule
\textbf{Ours} ($l_{\infty}\le9$) & 23.85 & \textbf{0.75} & \textbf{29.48} & \textbf{0.47} \\
\bottomrule
\end{tabular}
}
\caption{Comparison on image quality of adversarial examples with cross-model accuracy on ImageNet-1K.}
\label{tab:image_metrics}
\end{table}

\begin{figure}[!t]
\centering
\includegraphics[width=0.495\columnwidth]{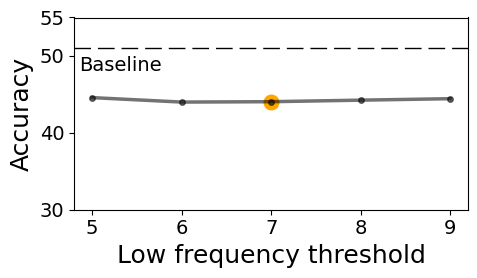}
\includegraphics[width=0.495\columnwidth]{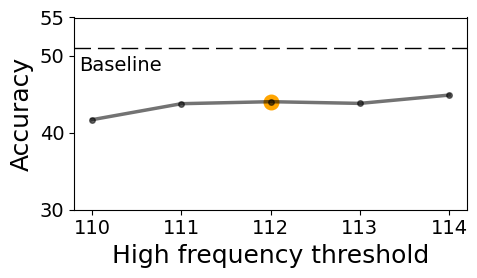}
\caption{Average cross-domain evaluation results across various frequency thresholds.}
\label{fig:freq_ranges}
\end{figure}

\subsubsection{Comparison with full-band randomization.}
We further investigated on the effectiveness of our domain randomization scheme, comparing with the full-band frequency randomization as practiced before~\cite{long2022frequency}.
As shown in Table~\ref{tab:freq_range_comparisons}, our novel domain-aware approach is superior to the naive full-range randomization method (\textit{i.e.}, \textit{All-Rand}).
Remarkably, \textit{All-Rand} is closely related to a recent work, namely SSA~\cite{long2022frequency}, which improves the iterative attack transferability by full-range spectral augmentation.
Compared to SSA, our method exclusively randomizes the \textit{domain-specific} low/high-FCs and exploits the frequency-augmented feature contrast.
Ours outperforms \textit{All-Rand} by $3.19\%$p and $2.02\%$p in each cross-domain and cross-model evaluation.
Without identifying and preserving \textit{domain-agnostic} information, even the state-of-the-art method could excessively randomize images, resulting in the degradation of image semantics and leading to the sub-optimal adversarial perturbation generation.

\subsubsection{Sensitivity on frequency thresholds.}
We investigated the sensitivity of the chosen frequency thresholds to verify the robustness of our approach.
As shown in Figure~\ref{fig:freq_ranges}, our method shows robust attack performance across adjacent threshold values, surpassing the baseline performance.
This implies that \textit{mid-frequency} range contains \textit{domain-agnostic} information that is effective in generating transferable perturbations against arbitrary domains and models.

\subsubsection{Analysis on image quality.}
\begin{figure}[!t]
\centering
\includegraphics[width=\linewidth]{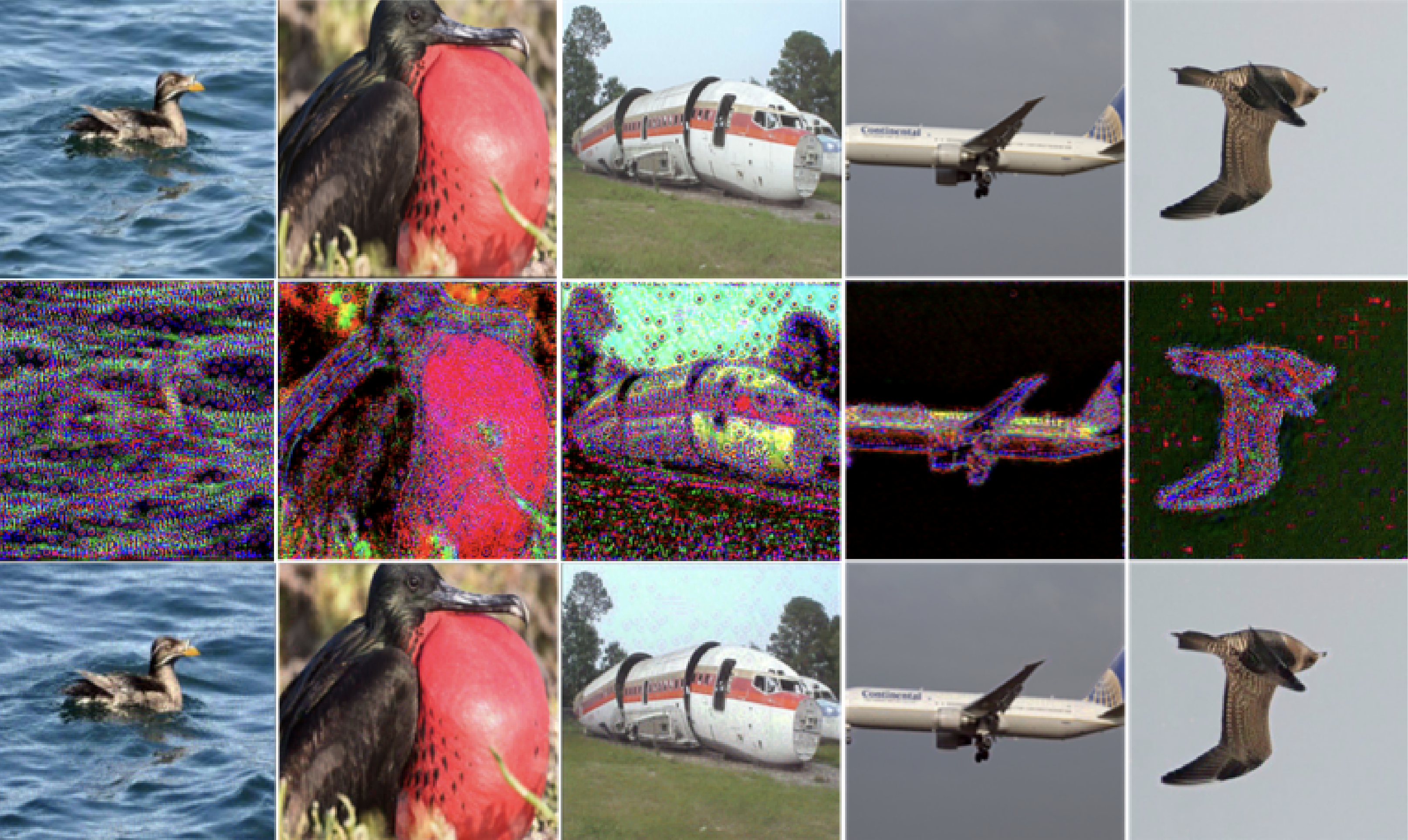}
\caption{Qualitative results. Clean images (row 1), unbounded adversarial images (row 2), and bounded ($l_\infty \leq 10$) adversarial images (row 3) are shown for various domains.
All of the final unbounded adversarial image samples cause victim classifier models to make incorrect predictions.
}
\label{fig:qualitative}
\end{figure}

Although our work is focused on generating more powerful adversarial perturbations, the image quality of the crafted adversarial examples should also be carefully examined.
As shown in Figure~\ref{fig:qualitative}, FACL-Attack can craft effective and high-quality adversarial images with imperceptible perturbations.
We also conducted a quantitative evaluation of image dissimilarity metrics between clean and adversarial image pairs, including SSIM, PSNR, and LPIPS.
As shown in Table~\ref{tab:image_metrics}, we found that ours with a lower perturbation of $l_{\infty}\le9$ demonstrates superior image quality than the baseline with $l_{\infty}\le10$ while achieving better attack performance.
In other words, it can yield better attack transferability with lower perturbation power and better image quality, which are very remarkable assets for real-world black-box attacks. We refer to \textit{Supplementary} for more qualitative and quantitative evaluation results.

\section{Conclusion}
In this paper, we have introduced a novel generator-based transferable attack method, leveraging spectral transformation and feature contrast in the frequency domain.
Our work drew inspiration from domain generalization approaches that utilize frequency domain techniques, adapting and enhancing them for the attack framework.
In our method, we target spectral data randomization on \textit{domain-specific} image components, and \textit{domain-agnostic} feature contrast for training a more robust perturbation generator.
Extensive evaluation results validate the effectiveness in practical black-box scenarios with domain shifts and model variances.
It can also be integrated into existing attack frameworks, further boosting the transferability while keeping the inference time.

\section{Acknowledgments}
This work was partially supported by the Agency for Defense Development grant funded by the Korean Government, and by the National Research Foundation of Korea (NRF) grant funded by the Korea government (MSIT) (NRF2022R1A2B5B03002636).
We thank Junhyeong Cho for his insightful discussions and valuable comments.


\clearpage
\appendix
\twocolumn[{
    \section*{\Huge{Supplementary Material}}
    \vspace{10mm}
}]
In this supplementary material, we provide contents that were not included in our main paper due to space limitations.
This includes additional experimental details, adapting FACL-Attack to other attacks, additional quantitative results, and additional qualitative results.
\vspace{3mm}

\section{Additional Experimental Details}
\label{sec:supp_experimental_details}
\setcounter{figure}{0}
\setcounter{table}{0}
\setcounter{algorithm}{0}
\setcounter{equation}{0}
\renewcommand{\thefigure}{A\arabic{figure}}
\renewcommand{\thetable}{A\arabic{table}}
\renewcommand{\thealgorithm}{A\arabic{algorithm}}
\renewcommand{\theequation}{A\arabic{equation}}

In this section, we provide additional experimental specifics on the algorithm details, implementation details, the FADR module, and the FACL module.

\subsection{Algorithm Details}
\label{sec:supp_algorithm_details}
We outline the algorithm details of FACL-Attack in Alg.~\ref{alg:FACL-Attack}.
The learning objective is to train a robust perturbation generator $G_\theta(\cdot)$ from which the crafted adversarial examples transfer well to unknown target domain regardless of data distributions or model architectures.
The training is entirely conducted in ImageNet-1K~\cite{imagenet} source domain with the data distribution of $\mathcal{X}_s$.

To elaborate on our training strategy, we first randomly initialize our perturbation generator $G_\theta(\cdot)$.
Next, we randomly sample a mini-batch ${\bm x}_s$ with batch size $N$, derived from the source data distribution $\mathcal{X}_s$.
To prevent excessive spectral transformation in $\mathcal{T}_{\mathrm{FADR}}(\cdot)$ and ensure stable training, we exclusively transform $N/2$ samples within the mini-batch in our FADR module.
The augmented samples $\tilde{\bm x}_s$ are then forward-passed through $G_\theta(\cdot)$ and the unbounded adversarial examples $\tilde{\bm x}'_s$ are crafted.
To ensure imperceptibility, the adversaries are constrained by the perturbation projection operator $P(\cdot)$.
Then, we forward-pass ${\bm x}'_s$ and $\tilde{\bm x}_s$ through the pre-trained surrogate model $\boldsymbol{f}$$_{k}(\cdot)$ after undergoing spectral decomposition $\mathcal{D}(\cdot)$.
Finally, we train the perturbation generator with the total loss objective that includes the baseline loss $\mathcal{L}_{\mathrm{orig}}$ and our contrastive loss $\mathcal{L}_{\mathrm{FACL}}$.

\subsection{Implementation Details}
\label{sec:supp_implementation_details}
Regarding the implementation of our generative attack, we adhere to the training pipeline and the generator architecture outlined in recent studies~\cite{poursaeed2018generative, naseer2019cross, salzmann2021learning, zhang2022beyond} for fair comparison.
Elaborating on the framework, a perturbation generator crafts an adversarial example from a clean input image, and the resulting unbounded adversarial example is constrained by a perturbation budget of $l_{\infty}\leq 10$.
Subsequently, the final pairs of adversarial and clean image are fed into the surrogate model for the attack.

For GAP~\cite{poursaeed2018generative}, we used their official code for the training.
As for CDA~\cite{naseer2019cross}, we also used their pre-trained models for evaluation.
We re-implemented LTP~\cite{salzmann2021learning}, utilizing the same generator architecture as BIA, but with their proposed $\mathcal{L}_{2}$ loss.
We set BIA~\cite{zhang2022beyond} as our baseline and implemented our proposed modules upon their code.
We conducted three different random runs to ensure reliability and reproducibility of our proposed method, which are reported in Table~\ref{tab:supp_multiple_random_runs}.
Note that our re-training of BIA (baseline) shows a slight better performance than the reported one in the paper.
The training takes around $14$ hours when using a single NVIDIA RTX A6000 GPU.
The S/W stack includes PyTorch 1.8.0, CUDA 11.1, and CUDNN 8.4.1.

\begin{algorithm}[!t]
    \small
    \caption{FACL-Attack}
    \label{alg:FACL-Attack}
    \begin{algorithmic}[1]
        \Require Source data distribution $\mathcal{X}_s$, surrogate model $f_{k}(\cdot)$, perturbation generator $G_\theta(\cdot)$, perturbation projector $P(\cdot)$, perturbation budget $\epsilon$, spectral transformation $\mathcal{T}_{\mathrm{FADR}}(\cdot)$, spectral decomposition $\mathcal{D}(\cdot)$
        \Ensure Randomly initialize the generator $G_\theta(\cdot)$
        \Repeat
        \State Randomly sample a mini-batch ${\bm x}_s \sim \mathcal{X}_s$ w/ batch size $N$
        \State Transform $\mathcal{T}_{\mathrm{FADR}}(\cdot)$ the $N/2$ samples in each mini-batch
        \State Prepare the augmented samples $\tilde{\bm x}_s = \mathcal{T}_{\mathrm{FADR}}({\bm x}_s)$
        \State Forward-pass $\tilde{\bm x}_s$ through $G_\theta(\cdot)$ and generate unbounded 
        \Statex \;\;\;\;\, adversarial examples $\tilde{\bm x}'_s$
        \State Bound the adversarial examples with $P(\cdot)$ such that:
        \begin{align*}
              \| P(\tilde{\bm x}'_s) - \tilde{\bm x}_s \|_{\infty} \le \epsilon \;\; \mathrm{and} \;\; P(\tilde{\bm x}'_s)={\bm x}'_s
        \end{align*}
        \State Forward-pass ${\bm x}'_s$ and $\tilde{\bm x}_s$ through $f_{k}(\cdot)$ for $\mathcal{L}_{\mathrm{orig}}$
        \State Forward-pass $\mathcal{D}({\bm x}'_s)$ and $\mathcal{D}(\tilde{\bm x}_s)$ through $f_{k}(\cdot)$ for $\mathcal{L}_{\mathrm{FACL}}$
        \State Compute the total loss $\mathcal{L}$
        \begin{align*}
              \mathcal{L} = \lambda_{\mathrm{orig}}\cdot\mathcal{L}_{\mathrm{orig}} + \lambda_{\mathrm{FACL}}\cdot\mathcal{L}_{\mathrm{FACL}}
        \end{align*}
        \State Backpropagate gradients and update $G_\theta(\cdot)$
        \Until{$G_\theta(\cdot)$ converges}
    \end{algorithmic}
\end{algorithm}

\begin{table}[!t]
\vspace{2mm}
\setlength{\tabcolsep}{11pt}
\centering
\resizebox{\linewidth}{!}{
\begin{tabular}{c|cc}
\toprule
Method & \multicolumn{1}{c}{Cross-Domain} & \multicolumn{1}{c}{Cross-Model}  \\
\midrule
\midrule
Baseline & 49.73 \small{$\pm$ 1.18} & 24.20 \small{$\pm$ 0.71} \\
+ FADR only & 46.24 \small{$\pm$ 0.21} & 20.28 \small{$\pm$ 1.25} \\
+ FACL only & 45.36 \small{$\pm$ 0.38} & 20.70 \small{$\pm$ 0.61} \\
+ FADR\;+\;FACL & 44.05 \small{$\pm$ 1.25} & 19.66 \small{$\pm$ 0.67} \\
\bottomrule
\end{tabular}
}
\caption{Multiple random runs with three different seeds.
We report the averaged top-1 classification accuracy after attacks (the lower, the better) with the standard deviation.
}
\label{tab:supp_multiple_random_runs}
\end{table}

\subsection{Details on FADR Module}
\label{sec:supp_FADR}
The objective of our FADR module is to convert a source-domain image $\bm x_s$ into an augmented sample $\tilde{\bm x}_s$ within the frequency domain.
This process is designed to to improve the training of the perturbation generator $G_\theta(\cdot)$ and thereby generate a more robust adversarial example with the same input dimensions.
Our spectral transformation $\mathcal{T}_{\mathrm{FADR}}(\cdot)$ randomizes domain-variant low- and high-frequency components (FCs) while keeping the domain-invariant mid-FCs. 
Building upon the insights from frequency threshold selection to segment frequency bands into low, mid, and high ranges as discussed in~\cite{huang2021fsdr}, we set the low and high frequency thresholds at $f_{l}=7$ and $f_{h}=112$ for an input image of dimensions $H\times W=224\times224$ after resizing. 
These thresholds have been adjusted proportionally in accordance with our input image size.

Our randomization scheme is closely related to the previous spectrum simulation attack (SSA)~\cite{long2022frequency}, which randomizes the Discrete Cosine Transform (DCT) converted frequency coefficients as a whole. In contrast, FACL-Attack is designed to specifically randomize the FCs in the domain-variant low- and high-frequency bands.
Moreover, SSA~\cite{long2022frequency} trains on a set of image samples augmented multiple times. 
In contrast, our approach applies randomization to only half of the samples within a mini-batch. This strategy is employed to enhance training stability and alleviate the risk of potential over-fitting.
Note that we use the whole ImageNet-1K~\cite{imagenet} ($\sim$128K) although they use ImageNet-compatible dataset ($\sim$1K) for the training dataset.

\begin{figure}[!t]
\centering
\includegraphics[width=0.49\columnwidth]{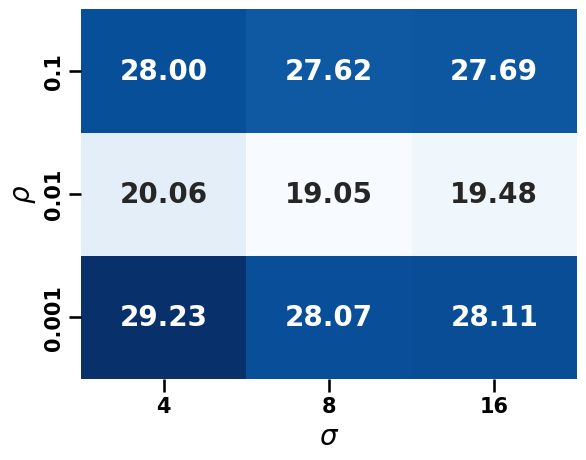}
\includegraphics[width=0.49\columnwidth]{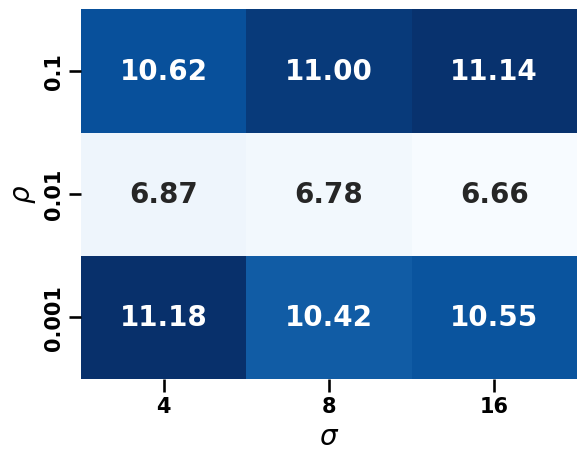}
\caption{
The averaged \textit{cross-domain} (left) and \textit{cross-model} (right) top-1 classification accuracy after attacks ($\downarrow$ is better) with respect to FADR hyperparameters of $\rho$ and $\sigma$.
}
\label{fig:supp_ablation_fadr}
\end{figure}

\begin{table}[!t]
\centering
\setlength{\tabcolsep}{2.5pt} 
\resizebox{\columnwidth}{!}{
\begin{tabular}{c|cccc|cc}
\toprule
Method & Low-Rand & Mid-Rand & High-Rand & All-Rand & FADR \\
\midrule
\midrule
Cross-Domain & 47.06 & 48.17 & 47.78 & 47.24 & \textbf{46.24} \\
Cross-Model & 21.25 & 22.99 & 22.79 & 21.68 & \textbf{20.28} \\
\bottomrule
\end{tabular}
}
\caption{Comparison with various band randomization.
We report the averaged top-1 classification after attacks (the lower, the better).}
\label{tab:supp_freq_range_comparisons}
\end{table}

\subsubsection{Hyperparameter selection.}
The spectral transformation operator $\mathcal{T}_{\mathrm{FADR}}(\cdot)$ has two hyperparameters: $\rho$ and $\sigma$. 
We conducted experiments to select an optimal combination of these hyperparameters, using Dense-169 as the surrogate model.
We used a different surrogate model from the one mentioned in the main paper (\textit{i.e.}, VGG-16) to ensure that the chosen randomization scheme is applicable in a broader context.
For the $\rho$, we vary the value from $0.001$ to $0.1$, increasing by a factor of ten. 
For the $\sigma$, we vary the value from $4$ to $16$, increasing by a factor of two. 
We evaluated the performance for each combination of $\rho$ and $\sigma$ in Figure~\ref{fig:supp_ablation_fadr}.
Given the significance of both of $\rho$ and $\sigma$, we set the combination of $\rho=0.01$ and  $\sigma=8$ as the optimal values for achieving superior improvements in the cross-domain setting. 
We speculate that excessive transformation could disturb the training of the generator.

\subsubsection{Comparison with various band randomization.}
We further investigated on the effectiveness of our domain randomization scheme with respect to each frequency band.
As shown in Table~\ref{tab:supp_freq_range_comparisons}, our novel frequency-aware randomization with domain knowledge is superior to other na\"ive band-specific or full-range randomization~\cite{long2022frequency} methods.
As \textit{Mid-Rand} degrades the performance compared to \textit{Low-Rand} and \textit{High-Rand}, we note that mid-band FCs contain more domain-invariant information to be preserved than domain-variant FCs in the low- or high-band.
Nonetheless, the overall performance is boosted compared to the baseline, and this could potentially be attributed to the Gaussian noise augmentation in our randomization module.

\subsection{Details on FACL Module}
\label{sec:supp_FACL}
The FACL module is designed to boost the robustness by leveraging the surrogate model $f(\cdot)$ to push apart the \textit{domain-invariant} mid-FCs feature pairs from clean and adversarial examples, while attract the \textit{domain-variant} low- and high-band FCs pairs each other. 
We have named this module ``frequency-augmented" since the contrasted feature pairs are augmented within the frequency domain before being fed into the surrogate model.
We use the same frequency thresholds as FADR for spectral decomposition $\mathcal{D}(\cdot)$, which is used to decompose the input images into mid-FCs and low-/high-FCs with band-pass and band-reject filters, respectively.
For implementing the baseline loss $\mathcal{L}_{\mathrm{orig}}$, we extract mid-layer features from \textit{Maxpool}.3 of VGG-16~\cite{vgg} as in BIA~\cite{zhang2022beyond}.
For implementing our contrastive loss $\mathcal{L}_{\mathrm{FACL}}$, we employ the $512$-dimensional mid-layer features (\textit{i.e.}, ReLU after \textit{Conv}4\_1) in line with the contrastive loss implementation of GAMA~\cite{aich2022gama}.

\begin{figure}[!t]
    \centering
    \includegraphics[width=0.99\linewidth,trim={0.1cm 0 0 8.3cm},clip,scale=1.00]{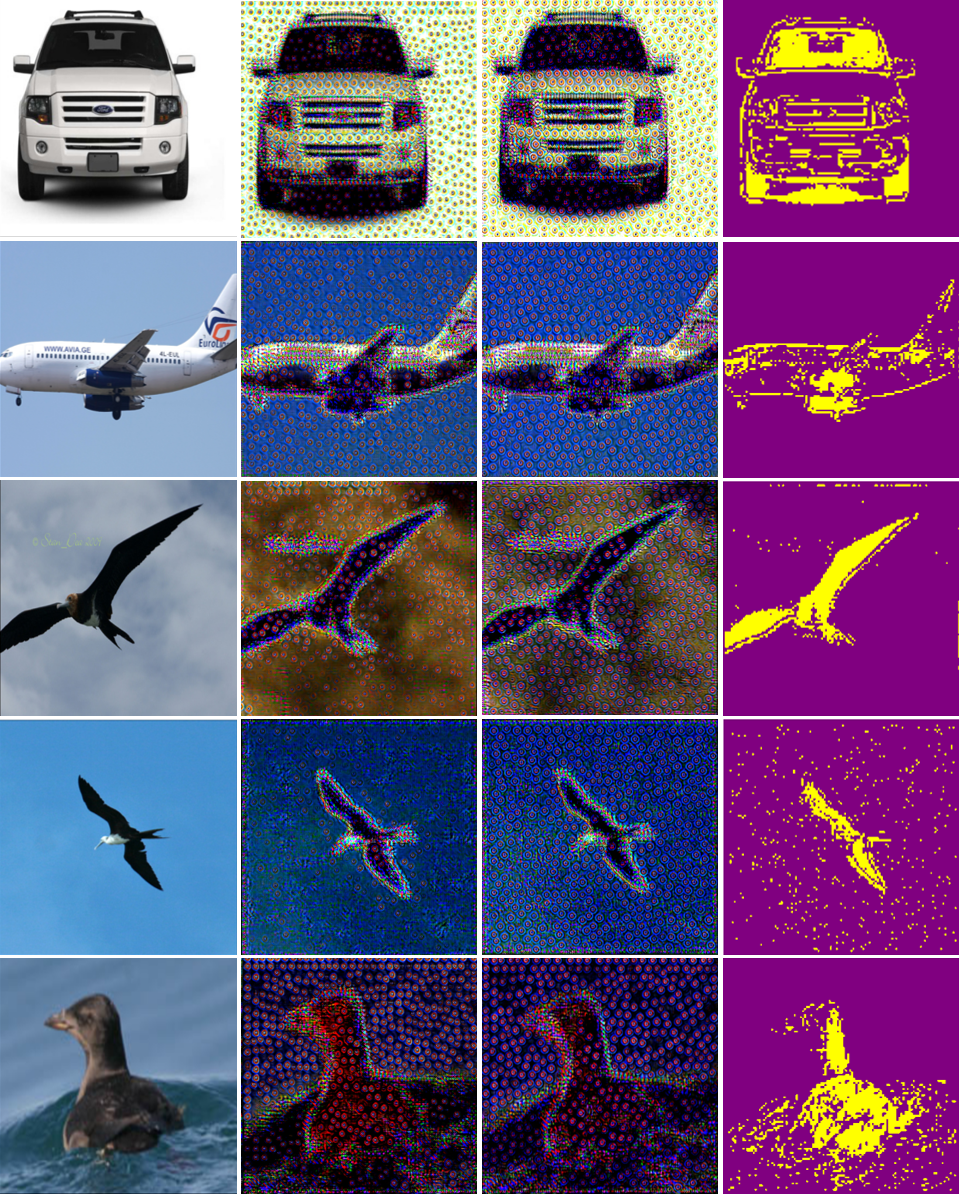}
    \caption{Clean image, unbounded adversarial image from baseline+FACL, and the difference map (\textit{Diff(baseline, baseline+FACL)}), from left to right.
    Our generated perturbations are more focused on domain-agnostic semantic region such as shape, facilitating more transferable attack.}
    \label{fig:supp_diff_map}
\end{figure}

\subsubsection{Difference map analysis.}
As discussed in BIA~\cite{zhang2022beyond}, we also utilize the difference map based on the generator features to conduct a more comprehensive analysis of the effectiveness and contributions of our proposed modules.
We employ a ResNet-based perturbation generator model architecture consisting of a series of down-sampling blocks, residual blocks, and up-sampling blocks.
Specifically, we take the output features of the down-sampling feature extraction block by applying cross-channel average pooling to obtain a difference map between the baseline and ours.
From the perspective of the generator feature space, we define the difference map as follows:
\begin{equation}
    Diff(F^{d}_{\mathrm{ours}}, F^{d}_{\mathrm{orig}}) = 
    \begin{cases}
        1, & F^{d}_{\mathrm{ours}} - F^{d}_{\mathrm{orig}} > 0,\\
        0, & else,
    \end{cases}
\label{eq:supp_diff_map}
\end{equation}
\noindent with $F^{d}_{\mathrm{orig}}$ and $F^{d}_{\mathrm{ours}}$ defined as:
\begin{align}
    F^{d}_{\mathrm{orig}} = \frac{| \sum_{k=0}^{C} G^{d(k)}_{\theta_\mathrm{orig}}(\bm x_{s}) |}{C}, \nonumber \\ 
    F^{d}_{\mathrm{ours}} = \frac{| \sum_{k=0}^{C} G^{d(k)}_{\theta_\mathrm{ours}}(\bm x_{s}) |}{C},
\label{eq:supp_generator_feat}
\end{align}
\noindent where $G^{d(k)}_{\theta_\mathrm{orig}}$ and $G^{d(k)}_{\theta_\mathrm{ours}}$ each denotes $k^{\mathrm{th}}$ channel output of the down-sampling block of baseline and ours, respectively.
In Figure~\ref{fig:supp_diff_map}, we present difference maps across various domains to compare our FACL with the baseline. It is noticeable that the object is more highlighted with our method compared to the baseline.
As this phenomenon is consistent across domains, we posit that our improved transferability could stem from the successful generation of perturbations in the domain-agnostic semantic region.

\begin{table}[!t]
\setlength{\tabcolsep}{17pt}
\centering
\resizebox{\linewidth}{!}{
\begin{tabular}{c|cc}
\toprule
\multirow{1}{*}{Method} & \multicolumn{1}{c}{Cross-Domain} & Cross-Model \\ 
\midrule
\midrule

\multicolumn{3}{c}{Adapting to BIA~\cite{zhang2022beyond}} \\
\midrule
BIA & 51.07 & 24.86 \\
BIA+$\mathcal{DA}$ & 40.65 & 19.60 \\
BIA+$\mathcal{RN}$ & 43.17 & 17.87 \\
\rowcolor{gray!20}Ours & 44.05 & 19.66 \\
\rowcolor{gray!20}Ours+$\mathcal{DA}$ & \textbf{38.46} & \textbf{16.93} \\
\rowcolor{gray!20}Ours+$\mathcal{RN}$ & 50.06 & 18.51 \\
\midrule

\multicolumn{3}{c}{Adapting to LTP~\cite{salzmann2021learning}} \\
\midrule
LTP & 49.91 & 22.30 \\
\rowcolor{gray!20}+Ours & \textbf{47.81} & \textbf{19.74} \\
\bottomrule
\end{tabular}
}
\caption{The averaged top-1 classification accuracy after attacks (the lower, the better), with adapting FACL-Attack to existing generative attacks.
The generator is trained on ImageNet-1K against VGG-16 surrogate model and evaluated on each black-box setting.
}
\label{tab:supp_adapt_bia}
\end{table}

\section{Adapting FACL-Attack to Other Attacks} 
\label{sec:supp_adapting}
\setcounter{figure}{0}
\setcounter{table}{0}
\setcounter{algorithm}{0}
\setcounter{equation}{0}
\renewcommand{\thefigure}{B\arabic{figure}}
\renewcommand{\thetable}{B\arabic{table}}
\renewcommand{\thealgorithm}{B\arabic{algorithm}}
\renewcommand{\theequation}{B\arabic{equation}}

To explore the versatility of our proposed modules with other existing generator-based methods, we conducted plug-and-play studies involving BIA~\cite{zhang2022beyond} and LTP~\cite{salzmann2021learning}.
With both our FADR and FACL modules in place, we evaluated the efficacy of our FACL-Attack by integrating it into the established training strategies, as depicted in Table~\ref{tab:supp_adapt_bia}.

\setcounter{figure}{0}
\setcounter{table}{0}
\setcounter{algorithm}{0}
\setcounter{equation}{0}
\renewcommand{\thefigure}{C\arabic{figure}}
\renewcommand{\thetable}{C\arabic{table}}
\renewcommand{\thealgorithm}{C\arabic{algorithm}}
\renewcommand{\theequation}{C\arabic{equation}}

\begin{table*}[!t]
\setlength{\tabcolsep}{3pt}
\centering
\resizebox{\linewidth}{!}{
    \begin{tabular}{ccccccccccccc}
        \toprule
        \multirow{2}{*}{\textbf{Method}} & \multicolumn{3}{c}{\textbf{CUB-200-2011}} && \multicolumn{3}{c}{\textbf{Stanford Cars}} && \multicolumn{3}{c}{\textbf{FGVC Aircraft}} & \multirow{2}{*}{\textbf{AVG.}} \\
        \cmidrule{2-12}
         
        & \textbf{Res-50} & \textbf{SENet154} & \textbf{SE-Res101} && \textbf{Res-50} & \textbf{SENet154} & \textbf{SE-Res101} &&  \textbf{Res-50} & \textbf{SENet154} & \textbf{SE-Res101} \\
        \midrule
        \midrule
        
        Clean & 87.35 & 86.81 & 86.56 && 94.35 & 93.36 & 92.97 && 92.23 & 92.08 & 91.90 & 90.85 \\
        \midrule
        
        PGD~\cite{madry2017towards} & 80.65 & 79.58 & 80.69 && 87.45 & 89.04 & 90.30 && 84.88 & 83.92 & 82.15 & 84.30\\
        
        DIM~\cite{DI} & 70.02 & 62.86 & \underline{70.57} && 74.72 & 78.10 & 84.33 && 73.54 & \underline{66.88} & \underline{62.38} & 71.49 \\
        
        DR~\cite{dr} & 81.08 & 82.05 & 82.52 && 90.82 & 90.59 & 91.12 && 84.97 & 87.55 & 85.54 & 86.25 \\
        
        SSP~\cite{ssp} & \underline{62.27} & \underline{60.44} & 71.52 && \underline{58.02} & \underline{75.71} & \underline{83.02} && \underline{54.91} & 68.74 & 63.79 & \underline{66.49} \\
        \midrule
        
        \textbf{FACL-Attack~(Ours)} & \textbf{24.74} & \textbf{44.06} & \textbf{53.75} && \textbf{26.58} & \textbf{65.71} & \textbf{61.40} && \textbf{19.72} & \textbf{52.01} & \textbf{48.51} & \textbf{44.05} \\                                     
        \bottomrule
    \end{tabular}
}
\caption{Comparison with iterative attacks. The perturbation generator is trained on ImageNet-1K against VGG-16 surrogate model and evaluated on black-box domains with black-box models.
We compare the top-1 classification accuracy after attacks with the perturbation budget of $l_\infty \leq 10$ (the lower, the better).
}
\label{tab:iterative_attacks}
\end{table*}

\begin{table*}[!t]
    \centering
    \setlength{\tabcolsep}{3pt}
    \resizebox{\linewidth}{!}
    {
    \begin{tabular}{ccccccccccccccccc}
        \toprule
         \multicolumn{1}{c}{\textbf{Method}} && \multicolumn{1}{c}{\textbf{WRN-50}} && \multicolumn{1}{c}{\textbf{MNasNet}} && \multicolumn{1}{c}{\textbf{MobileNetV3}} && \multicolumn{1}{c}{\textbf{ConvNeXt}} && \multicolumn{1}{c}{\textbf{ViT-B/16}} && \textbf{ViT-B/32} && \multicolumn{1}{c}{\textbf{ViT-L/16}} && \multicolumn{1}{c}{\textbf{ViT-L/32}} \\
        \midrule
        \midrule

        Clean && 77.24 && 66.49 && 73.09 && 83.93 && 79.56 && 76.91 && 80.86 && 76.52 \\
        \midrule
        
        GAP~\cite{poursaeed2018generative} && 59.72 && 42.47 && 56.54 && 79.68 && 72.89 && 71.10 && 76.69 && 71.40 \\
        
        CDA~\cite{naseer2019cross} && 35.85 && \underline{33.10} && 36.21 && \textbf{66.05} && 68.73 && 71.14 && 74.22 && 71.76 \\
        
        LTP~\cite{salzmann2021learning} && \textbf{22.66} && 45.28 && 43.30 && 70.43 && 72.44 && 72.69 && 76.75 && 72.73 \\
        
        BIA~\cite{zhang2022beyond} && 33.30 && 34.31 && \underline{35.26} && 69.17 && \underline{67.05} && \underline{68.15} && \underline{73.23} && \underline{69.78} \\
        \midrule
        
        \textbf{FACL-Attack (Ours)} && \underline{29.59} && \textbf{26.12} && \textbf{25.57} && \underline{67.17} && \textbf{65.21} && \textbf{64.82} && \textbf{71.48} && \textbf{67.25} \\
        
        \bottomrule
        \end{tabular}
    }
    \caption{Evaluation on the state-of-the-art models.
    The perturbation generator is trained on ImageNet-1K against VGG-16 surrogate model and evaluated on different network architectures.
    We compare the top-1 classification accuracy after attacks with the perturbation budget of $l_\infty \leq 10$ (the lower, the better).
    }
    \label{tab:eval_other_network_imagenet}
\end{table*}

\subsubsection{Adapting to BIA.}
Since we set BIA~\cite{zhang2022beyond} as our baseline, we already demonstrated the effectiveness of our proposed modules (\textit{i.e.}, FADR, FACL) when incorporated into BIA in the paper.
As there are two module variants in BIA (\textit{i.e.}, $\mathcal{DA}$ for domain-agnostic attention, and $\mathcal{RN}$ for random normalization), we conducted additional experiments with our FACL-Attack utilizing each BIA variant.
For the $\mathcal{DA}$, ``Ours+$\mathcal{DA}$" is superior to ``BIA+$\mathcal{DA}$", implying that our method could be compatible with $\mathcal{DA}$.
For the $\mathcal{RN}$, we conjecture that our FADR conflicts with the $\mathcal{RN}$ module, which additionally simulates different data distributions in the training pipeline.
We also note that $\mathcal{DA}$ and $\mathcal{RN}$ modules are not compatible together, as addressed in BIA~\cite{zhang2022beyond}.

\subsubsection{Adapting to LTP.}
We conducted another plug-and-play study on LTP~\cite{salzmann2021learning}, which leverages mid-level features of the surrogate model to learn an effective and transferable perturbation generator.
As shown in Table~\ref{tab:supp_adapt_bia}, our method can further enhance the attack transferability in both cross-domain and cross-model setting.

\section{Additional Quantitative Results}
\label{sec:supp_quantitative}
\setcounter{figure}{0}
\setcounter{table}{0}
\setcounter{algorithm}{0}
\setcounter{equation}{0}
\renewcommand{\thefigure}{C\arabic{figure}}
\renewcommand{\thetable}{C\arabic{table}}
\renewcommand{\thealgorithm}{C\arabic{algorithm}}
\renewcommand{\theequation}{C\arabic{equation}}

\begin{figure}[!t]
    \centering
    \includegraphics[width=\linewidth,trim={0 0 0 1cm},clip,scale=1.03]{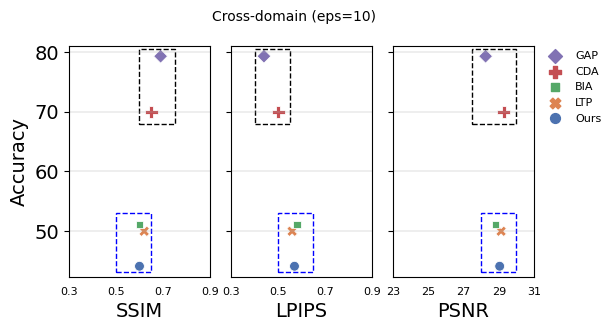}
    \caption{
    Comparison on the attack accuracy and the image quality metrics.
    The perturbation generator is trained on ImageNet-1K against VGG-16 surrogate model and evaluated on the various domains (\textit{i.e.}, CUB-200-2011, Stanford Cars, FGVC Aircraft).
    We report the averaged top-1 classification accuracy after attacks (the lower, the better) with a perturbation budget of $l_\infty \leq 10$.
    Ours achieves superior attack accuracy with competitive image quality scores.
    }
    \label{fig:cross_domain_eps10}
\end{figure}

\subsubsection{Attack accuracy and the image quality.}
We conducted additional quantitative evaluation on the image quality of the generated adversarial examples across domains in Figure~\ref{fig:cross_domain_eps10}.
Across the three perceptual metrics (\textit{i.e.}, SSIM, LPIPS, and PSNR), our method aligns with both LTP~\cite{salzmann2021learning} and BIA~\cite{zhang2022beyond} and does not exacerbate the image quality, while improving  the attack accuracy.
On the other hand, while there are slightly better quality scores with GAP~\cite{poursaeed2018generative} and CDA~\cite{naseer2019cross}, their attack performance falls below the expected standard.

\subsubsection{Comparison with iterative attacks.}
We compared our method against iterative-based adversarial attacks in Table~\ref{tab:iterative_attacks}.
The competitors include projected gradient descent (PGD)~\cite{madry2017towards}, diverse inputs method (DIM)~\cite{DI}, dispersion reduction (DR)~\cite{dr}, and self-supervised perturbation (SSP)~\cite{ssp}.
Following DR~\cite{dr}, we set the step size $\alpha=4$, and the number of iterations $T=100$ for all the iterative methods.
For DIM, we set the decay factor $\mu=1.0$ and the transformation probability $p=0.7$.
The results demonstrate the superiority of our novel generative method in terms of attack transferability.

\subsubsection{Evaluation on the state-of-the-art models.}
We report additional cross-model evaluation results on various state-of-the-art networks in Table~\ref{tab:eval_other_network_imagenet}, including WRN-50~\cite{zagoruyko2016wide}, MNasNet~\cite{tan2019mnasnet}, MobileNet V3~\cite{howard2019searching}, ConvNeXt~\cite{liu2022convnet}, and Vision Transformer (ViT)~\cite{dosovitskiy2020vit}.
Our method achieves superior cross-model transferability to other baselines.
Note that for the ViT, we used ImageNet-1K fine-tuned models that were pre-trained on ImageNet-21K.

\subsubsection{Training against other surrogate models.}
We included additional evaluation results with training against other surrogate models (\textit{i.e.},VGG-19 and Dense-169) on the final page: Table~\ref{tab:cross_domain_vgg_19}, \ref{tab:cross_domain_dense_169}, \ref{tab:cross_model_vgg_19}, and  \ref{tab:cross_model_dense_169}.
As shown in the results, our method consistently enhances the attack transferability across various settings, demonstrating the generalization capability of our approach.

\setcounter{figure}{0}
\setcounter{table}{0}
\setcounter{algorithm}{0}
\setcounter{equation}{0}
\renewcommand{\thefigure}{C\arabic{figure}}
\renewcommand{\thetable}{C\arabic{table}}
\renewcommand{\thealgorithm}{C\arabic{algorithm}}
\renewcommand{\theequation}{C\arabic{equation}}

\begin{table*}[!t]
\setlength{\tabcolsep}{3pt}
\centering
\resizebox{\linewidth}{!}{
    \begin{tabular}{ccccccccccccc}
        \toprule
        \multirow{2}{*}{\textbf{Method}} & \multicolumn{3}{c}{\textbf{CUB-200-2011}} && \multicolumn{3}{c}{\textbf{Stanford Cars}} && \multicolumn{3}{c}{\textbf{FGVC Aircraft}} & \multirow{2}{*}{\textbf{AVG.}} \\
        \cmidrule{2-12}
         
        & \textbf{Res-50} & \textbf{SENet154} & \textbf{SE-Res101} && \textbf{Res-50} & \textbf{SENet154} & \textbf{SE-Res101} &&  \textbf{Res-50} & \textbf{SENet154} & \textbf{SE-Res101} \\
        \midrule
        \midrule
        
        Clean & 87.35 & 86.81 & 86.56 && 94.35 & 93.36 & 92.97 && 92.23 & 92.08 & 91.90 & 90.85 \\
        \midrule
        
        GAP~\cite{poursaeed2018generative} & 77.39 & 77.29 & 77.34 && 87.30 & 87.48 & 88.27 && 79.45 & 80.86 & 76.36 & 81.30 \\
        
        CDA~\cite{naseer2019cross} & 59.48 & 61.08 & 68.50 && 58.53 & 70.70 & 80.70 && 59.26 & \textbf{52.24} & 62.26 &63.64\\
        
        LTP~\cite{salzmann2021learning} & \underline{42.70} & 55.09 & 68.59 && \textbf{37.74} & \textbf{68.44} & 80.54 && \underline{32.13} & 61.78 & 62.05 & \underline{56.56}\\
        
        BIA~\cite{zhang2022beyond} & 48.90 & \underline{52.33} & \underline{56.47} && 66.34 & 72.45 & \underline{75.08} && 50.95 & 54.04 & \underline{51.79} & 58.71 \\
        
        \midrule    
        
        \textbf{FACL-Attack~(Ours)} & \textbf{41.96} & \textbf{42.60} & \textbf{52.26} && \underline{37.78} & \underline{68.61} & \textbf{63.84} && \textbf{25.98} & \underline{53.11} & \textbf{44.64} & \textbf{47.86} \\                                    
        \bottomrule
    \end{tabular}
}
\caption{Cross-domain evaluation results. The perturbation generator is trained on ImageNet-1K with VGG-19 as the surrogate model and evaluated on black-box domains with black-box models.
We compare the top-1 classification accuracy after attacks with the perturbation budget of $l_\infty \leq 10$ (the lower, the better).
}
\label{tab:cross_domain_vgg_19}
\end{table*}

\begin{table*}[!t]
\setlength{\tabcolsep}{7pt}
\centering
\resizebox{\linewidth}{!}{
    \begin{tabular}{ccccccccccccccccccc}
        \toprule 
        \textbf{Method} &
        \textbf{Venue} &
        \multicolumn{1}{c}{\textbf{VGG-16}} &
        \multicolumn{1}{c}{\textbf{VGG-19}} &
        \multicolumn{1}{c}{\textbf{Res-50}} &
        \multicolumn{1}{c}{\textbf{Res-152}} &
        \multicolumn{1}{c}{\textbf{Dense-121}} &
        \multicolumn{1}{c}{\textbf{Dense-169}} &
        \multicolumn{1}{c}{\textbf{Inc-v3}} &                
        \multirow{1}{*}{\textbf{AVG.}} \\
        \midrule
        \midrule
        
        Clean & - & 70.14 & 70.95 & 74.61 & 77.34 & 74.22 & 75.75 & 76.19 & 74.17 \\         
        \midrule
        
        GAP~\cite{poursaeed2018generative} & CVPR'18 & 36.56 & 29.44 & 61.10 & 67.49 & 60.77 & 64.69 & 65.50 & 55.08\\
        
        CDA~\cite{naseer2019cross} & NeurIPS'19 & \textbf{1.09} & \textbf{0.26} & \underline{24.95} & 44.64 & 39.00 & 42.97 & 55.22 & 29.73\\
        
        LTP~\cite{salzmann2021learning} & NeurIPS'21 & 2.40 & 1.84 & \textbf{21.61} & \textbf{41.17} & 30.09 & \textbf{31.87}  &  \underline{46.39} & \underline{25.05}\\
        
        BIA~\cite{zhang2022beyond} & ICLR'22 & 
        2.50 & 1.88 & 25.40 & \underline{41.60} & \underline{29.81} & 37.08 & 46.59 & 26.41 \\ 
        \midrule
        
        \textbf{FACL-Attack~(Ours)} & - & \underline{2.07} & \underline{1.18} & 25.40 & 44.07 & \textbf{29.01} & \underline{34.00} & \textbf{34.17} & \textbf{24.27} \\
        \bottomrule
         
    \end{tabular}
}
\caption{Cross-model evaluation results. The perturbation generator is trained on ImageNet-1K against VGG-19 surrogate model and evaluated on black-box models including white-box model (\textit{i.e.}, VGG-19).
We compare the top-1 classification accuracy after attacks with the perturbation budget of $l_\infty \leq 10$ (the lower, the better).
}
\label{tab:cross_model_vgg_19}
\end{table*}

\begin{table*}[!t]
\setlength{\tabcolsep}{3pt}
\centering
\resizebox{\linewidth}{!}{
    \begin{tabular}{ccccccccccccc}
        \toprule
        \multirow{2}{*}{\textbf{Method}} & \multicolumn{3}{c}{\textbf{CUB-200-2011}} && \multicolumn{3}{c}{\textbf{Stanford Cars}} && \multicolumn{3}{c}{\textbf{FGVC Aircraft}} & \multirow{2}{*}{\textbf{AVG.}} \\
        \cmidrule{2-12}
         
        & \textbf{Res-50} & \textbf{SENet154} & \textbf{SE-Res101} && \textbf{Res-50} & \textbf{SENet154} & \textbf{SE-Res101} &&  \textbf{Res-50} & \textbf{SENet154} & \textbf{SE-Res101} \\
        \midrule
        \midrule
        
        Clean & 87.35 & 86.81 & 86.56 && 94.35 & 93.36 & 92.97 && 92.23 & 92.08 & 91.90 & 90.85 \\
        \midrule
        
        GAP~\cite{poursaeed2018generative} & 60.87 & 72.39 & 68.17 && 77.63 & 83.72 & 84.84 && 75.46 & 80.02 & 72.64 & 75.08\\ 
        
        CDA~\cite{naseer2019cross} & 52.92 & 60.96 & 57.04 && 53.64 & 73.66 & 75.51 && 62.23 & 61.42 & 59.83 &61.91\\
        
        LTP~\cite{salzmann2021learning} & \underline{19.97} & 34.09 & 45.48 && \underline{4.81}  & 47.61 & \underline{46.05} && \underline{5.19}  & \underline{19.71} & \underline{26.16} & \underline{27.67}\\
        
        BIA~\cite{zhang2022beyond} & 21.79 & \underline{29.29} & \underline{39.13} && 9.58 & \underline{44.46} & 49.06 && 8.04 & 27.84 & 33.87 & 29.23 \\
        
        \midrule    
        
        \textbf{FACL-Attack~(Ours)} & \textbf{9.65}  & \textbf{28.13} & \textbf{33.71} && \textbf{4.04}  & \textbf{39.07} & \textbf{25.87} && \textbf{3.54}  & \textbf{14.67} & \textbf{12.78} & \textbf{19.05} \\                                  
        \bottomrule
    \end{tabular}
}
\caption{Cross-domain evaluation results. The perturbation generator is trained on ImageNet-1K with Dense-169 as the surrogate model and evaluated on black-box domains with black-box models.
We compare the top-1 classification accuracy after attacks with the perturbation budget of $l_\infty \leq 10$ (the lower, the better).
}
\label{tab:cross_domain_dense_169}
\end{table*}

\begin{table*}[!t]
\setlength{\tabcolsep}{7pt}
\centering
\resizebox{\linewidth}{!}{
    \begin{tabular}{ccccccccccccccccccc}
        \toprule 
        \textbf{Method} &
        \textbf{Venue} &
        \multicolumn{1}{c}{\textbf{VGG-16}} &
        \multicolumn{1}{c}{\textbf{VGG-19}} &
        \multicolumn{1}{c}{\textbf{Res-50}} &
        \multicolumn{1}{c}{\textbf{Res-152}} &
        \multicolumn{1}{c}{\textbf{Dense-121}} &
        \multicolumn{1}{c}{\textbf{Dense-169}} &
        \multicolumn{1}{c}{\textbf{Inc-v3}} &                
        \multirow{1}{*}{\textbf{AVG.}} \\
        \midrule
        \midrule
        
        Clean & - & 70.14 & 70.95 & 74.61 & 77.34 & 74.22 & 75.75 & 76.19 & 74.17 \\         
        \midrule
         
        GAP~\cite{poursaeed2018generative} & CVPR'18 & 39.11 & 39.62 & 50.72 & 58.33 & 49.04 & 42.67 & 48.08 & 46.80 \\
          
        CDA~\cite{naseer2019cross} & NeurIPS'19 & 7.26 & 7.91 & 6.46 & 15.56 & \underline{5.13} & \textbf{0.63} & 43.78  & 12.39 \\
        
        LTP~\cite{salzmann2021learning} & NeurIPS'21 & 5.93 & 7.52  & \underline{6.34} & \underline{10.73} & 6.68  & 4.39 & 40.92 & \underline{11.79}\\
        
        BIA~\cite{zhang2022beyond} & ICLR'22 & \underline{4.76} & \underline{7.15} & 6.97 & 13.83 & 6.60& 6.45 & \underline{38.58} & 12.05 \\
        \midrule
        
        \textbf{FACL-Attack~(Ours)} & - & \textbf{2.78} & \textbf{3.68} & \textbf{3.78}  & \textbf{5.07}  & \textbf{3.56} & \underline{2.84}  & \textbf{25.74} & \textbf{6.78} \\
        \bottomrule
         
    \end{tabular}
}
\caption{Cross-model evaluation results. The perturbation generator is trained on ImageNet-1K with Dense-169 as the surrogate model and evaluated on black-box models including white-box model (\textit{i.e.}, Dense-169).
We compare the top-1 classification accuracy after attacks with the perturbation budget of $l_\infty \leq 10$ (the lower, the better).
}
\label{tab:cross_model_dense_169}
\end{table*}

\section{Additional Qualitative Results}
\label{sec:supp_qualitative}
\setcounter{figure}{0}
\setcounter{table}{0}
\setcounter{algorithm}{0}
\setcounter{equation}{0}
\renewcommand{\thefigure}{D\arabic{figure}}
\renewcommand{\thetable}{D\arabic{table}}
\renewcommand{\thealgorithm}{D\arabic{algorithm}}
\renewcommand{\theequation}{D\arabic{equation}}

\begin{figure*}[!t]
    \centering
    \includegraphics[width=\linewidth]{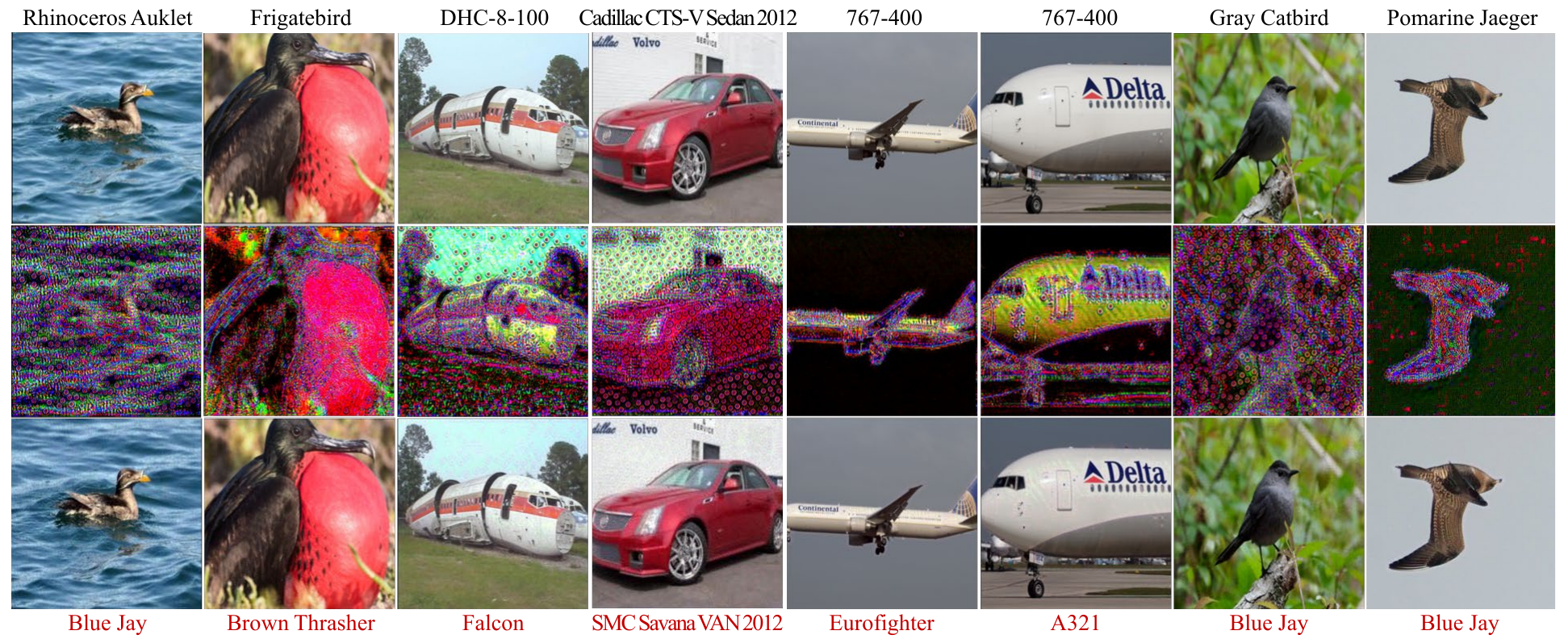}
    \caption{Additional qualitative results. Clean images (\textit{row 1}), unbounded adversarial images (\textit{row 2}), and bounded adversarial images (\textit{row 3}; actual inputs to the classifier) are shown for various domains.
    The ground truth and each mis-predicted class label are shown on the \textit{top} and \textit{bottom}.}
    \label{fig:supp_fine_adv_samples}
\end{figure*}

We show our generated adversarial samples crafted using FACL-Attack in Figure~\ref{fig:supp_fine_adv_samples} for multiple datasets, including CUB-200-2011~\cite{cub}, Stanford Cars~\cite{car}, and FGVC Aircraft~\cite{air}.
As evident from the visualization of unbounded adversarial examples, FACL-Attack encourages the generator to focus more on the object itself.
This phenomenon becomes more noticeable when the background color is uniform and solid.
For the unbounded adversarial examples in the middle row with the ground truth class labels ``767-400" and ``Pomarine Jaeger," the perturbations are concentrated more on the domain-agnostic semantic region, such as the object's shape.
While the visually displayed unbounded adversarial examples seem to undergo significant transformations, the resulting bounded examples maintain an almost imperceptible level of visual distortion, adhering to a perturbation budget of $l_{\infty}\leq10$.
Most importantly, our generated adversarial images are successful in inducing misclassification in the unknown victim models and domains.

\end{document}